%% file: iclr2027_conference.tex
\documentclass{article} 
\usepackage{iclr2027_conference,times}
\usepackage{booktabs}
\usepackage{multirow}
\usepackage{graphicx}
\usepackage{algorithm}
\usepackage{algpseudocode}
\input{math_commands.tex}

\usepackage{hyperref}
\usepackage{url}

\title{Beyond Full-Context Accuracy: Context-Marginalized Evaluation of Test-Time Structure-Aware Inference in EEG-to-Image Retrieval}

\author{
 \makebox[\dimexpr\textwidth-2\tabcolsep\relax][c]{%
    \begin{tabular}[t]{@{}c@{}}
      {\large\bfseries Ke Xiao, Banghua Yang, Yingjie Li, Hang Zhang}\\[4pt]
      {\normalfont Shanghai University, Shanghai, China}\\
      {\normalfont\ttfamily xiaoke77@shu.edu.cn}
    \end{tabular}%
  }%
}

\iclrfinalcopy 
\begin{document}

\maketitle
\fancyhead{}
\renewcommand{\headrulewidth}{0pt}
\begin{abstract}
Test-time structure-aware inference can improve cross-subject EEG-to-image retrieval by exploiting relations among unlabeled test queries, but it also makes predictions depend on query-set composition. Existing evaluations typically report accuracy under a single full-query context, leaving unclear whether gains and comparative conclusions remain reliable when the available context changes. We introduce Context-Marginalized Retrieval Evaluation (CMRE), which repeatedly evaluates each target query within sampled subsets of companion queries, applying each inference rule to the corresponding rows of a fixed query–gallery similarity matrix. CMRE distinguishes structural exploitability, quantified by gains over independent retrieval, from context robustness, measured by same-query Top-1 context consistency. Under leave-one-subject-out evaluation on THINGS-EEG2 across six frozen representations and six structure-aware inference rules, structure-aware inference improves full-context Top-1 accuracy by up to 20.40 percentage points over independent retrieval. However, the rule selected under full-context evaluation is no longer optimal in 17 of 24 representation--context-size conditions (70.8\%). Selecting the full-context winner incurs an average loss of 2.70 percentage points and a maximum loss of 8.29 percentage points under reduced contexts. Representation rankings are likewise inference- and context-dependent: the ordering under independent retrieval is not consistently preserved under structure-aware inference. Moreover, larger retrieval gains do not necessarily imply greater context robustness, showing that retrieval effectiveness and prediction stability provide complementary views of structure-aware inference. Together, these results show that full-context accuracy represents only one operating point of context-dependent inference. CMRE complements conventional evaluation by characterizing structural gain, context consistency, and the context dependence of conclusions about inference rules and frozen EEG-to-image representations.

\end{abstract}

\section{Introduction}
\label{sec:introduction}
EEG-to-image retrieval seeks to identify visually corresponding images in a candidate gallery by aligning EEG representations with visual embedding spaces. Recent advances in neural representation learning and cross-modal alignment have substantially improved retrieval performance \citep{song2024decoding,wu2025ubp,zhang2025cognitioncapturer,li2024visual,zheng2026brainhive,jo2026hyfi}.  Despite this progress, cross-subject generalization remains challenging because of pronounced inter-subject variability in EEG signals. This challenge has motivated leave-one-subject-out (LOSO) evaluation, in which models must generalize to previously unseen subjects without subject-specific supervised retraining. Beyond representation learning, recent work has begun to exploit information from unlabeled target-subject queries at test time to improve cross-subject retrieval without labeled calibration \citep{huang2026sattc,cui2026score}. However, once test-time inference exploits relations among multiple target queries, retrieval becomes context-dependent: the prediction for the same EEG query may change with the composition of the other queries processed alongside it.

This coupling creates an evaluation problem. For a fixed gallery, independent retrieval predicts each query from its own similarity vector and is therefore unaffected by the other test queries. By contrast, inference rules based on set-level statistics or global assignment constraints couple each prediction to the remaining test queries. Their measured performance therefore reflects both the quality of the frozen representation and the particular query context used for evaluation. A single full-context operating point can consequently conceal sensitivity to query-set composition. Existing cross-subject EEG-to-image studies that exploit unlabeled target-side information at test time do not systematically examine how retrieval performance and same-query prediction stability vary with query context \citep{huang2026sattc,cui2026score}. Related sensitivity to the composition and size of unlabeled test sets has also been reported in transductive learning \citep{veilleux2021realistic,martin2022practical,yuan2023robust}. These observations motivate two complementary questions. First, how much additional retrieval performance can structure-aware inference extract from a fixed similarity matrix? Second, how stable are the predictions for the same EEG query when the available companion-query context changes? We refer to these properties as \textbf{structural exploitability} and \textbf{context robustness}, respectively. Beyond these two properties, we further examine whether comparative conclusions about inference rules and frozen EEG-to-image representations remain stable across test-time inference conditions.

To address these questions, we introduce Context-Marginalized
Retrieval Evaluation (CMRE). CMRE repeatedly evaluates each target query across sampled query contexts at predefined sizes while ensuring that every target is evaluated equally often. For each frozen representation, all inference rules operate on context-specific rows drawn from the same full query--gallery similarity matrix. CMRE reports context-marginalized retrieval performance, quantifies structural exploitability through gains over independent retrieval, and measures context robustness through the consistency of same-query Top-1 predictions across contexts. Under LOSO cross-subject protocols on THINGS-EEG2~\citep{gifford2022things}, we apply CMRE to six frozen EEG-to-image representations and six structure-aware inference rules spanning set-level and assignment-based inference, with independent retrieval serving as the context-invariant reference. Within each frozen representation, this controlled design isolates differences attributable to test-time inference by providing all inference rules with identical query--gallery similarity evidence.

Our results reveal several limitations of relying on a single full-context evaluation. Across six frozen EEG-to-image representations and six structure-aware inference rules, test-time structure improves full-context Top-1 accuracy by up to 20.40 percentage points over independent retrieval. However, the inference rule ranked first under full-context evaluation is no longer optimal in 17 of 24 representation--context-size conditions (70.8\%). Selecting the full-context winner incurs an average performance loss of 2.70 percentage points and a maximum loss of 8.29 percentage points under reduced context. The apparent ranking of frozen representations is also inference- and context-dependent: the ordering observed under independent retrieval is not consistently preserved under structure-aware inference. Moreover, larger gains over independent retrieval do not necessarily coincide with greater same-query context consistency, showing that retrieval effectiveness and context robustness capture distinct properties of structure-aware inference. Together, these findings indicate that full-context accuracy represents only one operating point of context-dependent inference and can provide an incomplete basis for comparing both inference strategies and frozen EEG-to-image representations. Our main contributions are summarized as follows:

\begin{itemize}

\item We formalize test-time EEG-to-image retrieval under fixed
query--gallery similarity scores and distinguish three inference regimes: independent retrieval, set-level inference, and assignment-based inference. This formulation enables controlled comparisons of inference rules while holding the underlying representation fixed.

\item We identify \textbf{structural exploitability} and
\textbf{context robustness} as two complementary properties of test-time structure-aware inference and introduce \textbf{Context-Marginalized Retrieval Evaluation (CMRE)} to quantify both. CMRE repeatedly evaluates each target query under randomly sampled query contexts of predefined sizes while balancing target-query evaluation frequency. It measures gains over independent retrieval and same-query Top-1 consistency across contexts.

\item Through LOSO cross-subject evaluation across six
frozen EEG-to-image representations and six structure-aware inference rules, we show that larger structural gains do not necessarily imply greater context robustness. More importantly, the inference rule ranked first under full-context evaluation is no longer optimal in 17 of 24 representation--context-size conditions (70.8\%), incurring an average performance loss of 2.70 percentage points and a maximum loss of 8.29 percentage points under reduced context. We further show that the apparent ranking and relative performance margins of frozen representations depend on how test-time structure is exploited and can vary with the available query context.

\end{itemize}

\section{Problem Formulation}

\label{subsec:retrieval_regimes}
\begin{figure*}[t]
    \centering
    \includegraphics[width=\textwidth]{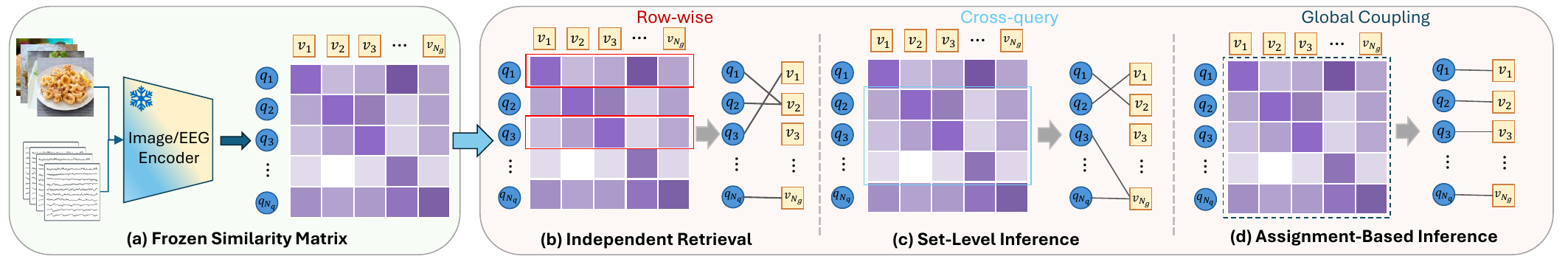}
    \caption{
    Overview of the frozen-similarity retrieval setting and the three
    test-time inference regimes considered in this work.
    (a) Frozen EEG--image representations are used to construct the similarity matrix $\mathbf{S}$.
    (b) Independent retrieval performs row-wise ranking without
    cross-query interaction.
    (c) Set-level inference exploits cross-query statistics to recalibrate similarity scores while retaining query-wise prediction.
    (d) Assignment-based inference jointly couples query--gallery
    correspondences through global coupling.
    }
    \label{fig:inference_regimes}
\end{figure*}

Let $\mathcal{Q}=\{q_i\}_{i=1}^{N_q}$ denote the EEG queries and
$\mathcal{G}=\{g_j\}_{j=1}^{N_g}$ the image gallery. Frozen EEG and image
encoders produce normalized embeddings $\mathbf{z}_i$ and $\mathbf{v}_j$,
from which we construct the query--gallery similarity matrix
\begin{equation}
\mathbf{S}\in\mathbb{R}^{N_q\times N_g},
\qquad
S_{ij}=\mathbf{z}_i^{\top}\mathbf{v}_j .
\end{equation}
All inference regimes considered in this work operate on the same frozen
similarity matrix and differ only in how they use information across queries,
as illustrated in Fig.~\ref{fig:inference_regimes}.

Under independent retrieval (Fig.~\ref{fig:inference_regimes}(b)), each query is
predicted using only its own similarity row:
\begin{equation}
\hat{y}_i^{\mathrm{ind}}
=
\operatorname*{arg\,max}_{j} S_{ij}.
\end{equation}
Because $\hat{y}_i^{\mathrm{ind}}$ depends only on
$\mathbf{S}_{i,:}$, it is invariant to the presence or composition of the
other test queries. Let $\mathcal{B}\subseteq\mathcal{Q}$, with $|\mathcal{B}|=m$, denote a query
context processed jointly at inference. A context-dependent inference rule
$\Phi$ operates on the corresponding rows of the frozen similarity matrix:
\begin{equation}
\hat{\mathbf{y}}^{(\mathcal{B})}
=
\Phi\!\left(\mathbf{S}_{\mathcal{B},:}\right).
\end{equation}
For two contexts $\mathcal{B}$ and $\mathcal{B}'$ that both contain $q_i$,
the resulting prediction for $q_i$ may differ even though its similarity row,
the image gallery, and the underlying representations remain fixed.

As illustrated in Fig.~\ref{fig:inference_regimes}(c,d), set-level inference uses
cross-query statistics to recalibrate similarity scores while retaining
query-wise prediction, whereas assignment-based inference jointly couples
query--gallery correspondences. This distinction describes how cross-query
information enters inference rather than an ordering of method strength;
stronger coupling does not necessarily yield better retrieval performance.

\section{Context-Marginalized Retrieval Evaluation}
\label{sec:cmre}

\begin{figure*}[t]
    \centering
    \includegraphics[width=\textwidth,height=3cm]{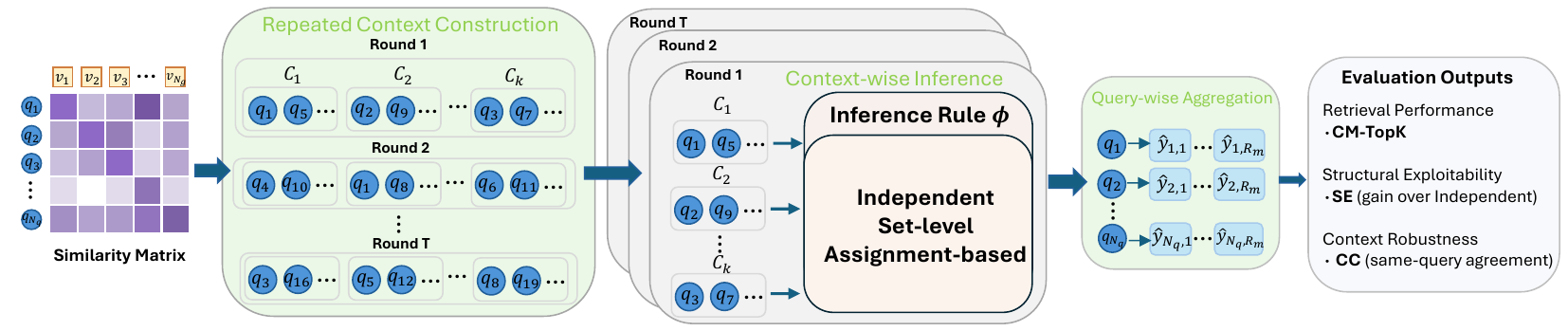}
   \caption{
Overview of Context-Marginalized Retrieval Evaluation (CMRE). Repeated
context construction and context-wise inference generate per-query
observations from a fixed query--gallery similarity matrix; query-wise
aggregation yields CM-TopK, structural exploitability relative to independent
retrieval, and same-query Context Consistency (CC).
}
    \label{fig:cmre}
\end{figure*}

A single full-context evaluation captures only one operating point of context-dependent inference. We therefore introduce
\textbf{Context-Marginalized Retrieval Evaluation (CMRE)}, which marginalizes evaluation over repeatedly constructed query contexts while keeping the full query--gallery similarity matrix fixed (Fig.~\ref{fig:cmre}). Each inference
rule is applied independently to every context, and the resulting per-query predictions and ranks are aggregated across context evaluations. CMRE reports context-marginalized retrieval performance, quantifies \textbf{structural exploitability} (SE) through gains over independent retrieval, and measures \textbf{context robustness} using Context Consistency (CC), the pairwise agreement of each query's Top-1 predictions across contexts.

\paragraph{Repeated Context Construction.}
For a context size $m$, CMRE repeatedly constructs query contexts at the test-set level according to a balanced evaluation schedule, and every query contained in a constructed context is evaluated. For a given query $q_i$, let
$\mathcal{B}_{i,r}^{(m)}$ denote the $r$-th constructed context containing
$q_i$, where $r\in\{1,\ldots,R_m\}$ and
\begin{equation}
q_i\in\mathcal{B}_{i,r}^{(m)}, \qquad
\left|\mathcal{B}_{i,r}^{(m)}\right|=m.
\end{equation}

We evaluate reduced query-context sizes $m\in\{25,50,100,150\}$ using $T=10$ construction rounds. For $m\in\{25,50,100\}$, the $N_q=200$ test queries are randomly permuted in each round and partitioned into $N_q/m$ non-overlapping groups of size $m$, denoted by $\{C_k\}_{k=1}^{N_q/m}$. Thus, every query appears in exactly one context per round. For $m=150$, the queries are randomly partitioned into four disjoint groups of 50 in each round, and four 150-query contexts, denoted by $\{C_k\}_{k=1}^{4}$, are constructed as their complements. Every query therefore appears in three contexts per round. Consequently, each query receives $R_{25}=R_{50}=R_{100}=10$ evaluations and $R_{150}=30$ evaluations. The schedule equalizes the number of evaluations per query at each context size, while leaving pairwise companion-query co-occurrence unconstrained. The conventional full-context case, $m=N_q=200$, is evaluated separately. The complete CMRE evaluation procedure and context-sampling schedule are provided in Appendix~\ref{app:cmre_algorithm}.

\paragraph{Context-Marginalized Retrieval Performance.}

Let $y_i$ denote the index of the ground-truth image for query $q_i$, and let
$\rho_{i,r}^{(m,\Phi)}$ denote its rank under inference rule $\Phi$ in the
$r$-th context evaluation at context size $m$. We define
\begin{equation}
\operatorname{CM\text{-}TopK}_{\Phi}(m)
=
\frac{1}{N_qR_m}
\sum_{i=1}^{N_q}
\sum_{r=1}^{R_m}
\mathbb{I}
\left[
\rho_{i,r}^{(m,\Phi)}\leq K
\right],
\end{equation}
Setting $K=1$ yields
$\operatorname{CM\text{-}Top1}_{\Phi}(m)$. This metric averages Top-$K$
retrieval success across all query--context evaluations. Let $\rho_i^{\mathrm{ind}}$ denote the rank of the ground-truth image for
query $q_i$ under independent retrieval. The independent-retrieval reference
at rank $K$ is
\begin{equation}
\mathrm{TopK}_{\mathrm{ind}}
=
\frac{1}{N_q}
\sum_{i=1}^{N_q}
\mathbb{I}\!\left[\rho_i^{\mathrm{ind}} \le K\right].
\end{equation}

We quantify SE as the gain over independent retrieval at the corresponding rank cutoff $K$:
\begin{equation}
\mathrm{SE}_{\Phi}(m)
=
\mathrm{CM\text{-}TopK}_{\Phi}(m)
-
\mathrm{TopK}_{\mathrm{ind}}.
\end{equation}
A positive value indicates that the inference rule benefits from cross-query structure relative to independent query-wise retrieval. For the full-context case, we set $R_{N_q}=1$, such that $\mathrm{CM\text{-}TopK}_{\Phi}(N_q)$ reduces to conventional full-context Top-$K$ accuracy. The same definition therefore applies to both reduced and full contexts. 

\textbf{Same-Query Context Consistency.}
Context-marginalized retrieval performance does not indicate whether the same query receives stable Top-1 predictions across contexts. We therefore define CC. Let $n_{i,c}^{(m,\Phi)}$ denote the number of context evaluations in which inference rule $\Phi$ predicts gallery candidate $c$ as the Top-1 match for query $q_i$. We define
\begin{equation}
\mathrm{CC}_{\Phi}(m)
=
\frac{1}{N_q}
\sum_{i=1}^{N_q}
\frac{
\sum_{c=1}^{N_g}
n_{i,c}^{(m,\Phi)}
\left(n_{i,c}^{(m,\Phi)}-1\right)
}{
R_m(R_m-1)
}.
\end{equation}
For each query, CC measures the fraction of pairs of distinct context evaluations that yield the same Top-1 prediction, and then averages this agreement across queries. It therefore measures context robustness rather than retrieval correctness. Independent retrieval is context-invariant and therefore attains
$\operatorname{CC}_{\mathrm{ind}}(m)=100\%$.
In the full-context case, each query is evaluated only once
($R_{N_q}=1$), so CC cannot be computed and is therefore not reported for
$m=N_q$.

\section{Experiments}
\label{sec:experiments}

\subsection{Experimental Setup}
\label{subsec:experimental_setup}

\paragraph{Dataset and Retrieval Protocol.}

We evaluate THINGS-EEG2~\citep{gifford2022things} under the standard 200-way EEG-to-image retrieval setting using a leave-one-subject-out (LOSO) cross-subject protocol, with results averaged across ten held-out subjects.  We additionally perform cross-modal validation on THINGS-MEG ~\citep{hebart2023things} under a LOSO protocol across four held-out subjects.

\paragraph{Representations.}
We evaluate six frozen EEG-to-image representations: NICE \citep{song2024decoding}, ATM \citep{li2024visual}, UBP \citep{wu2025ubp}, BLUR \citep{liu2026leveraging}, HyFI \citep{jo2026hyfi}, and BrainHIVE \citep{zheng2026brainhive}. These methods cover diverse neural--visual alignment strategies and representation strengths, allowing us to assess whether test-time structure-aware inference behaves consistently across different EEG--image similarity spaces.

\paragraph{Compared Inference Methods.}
We compare independent retrieval with three set-level methods: CSLS \citep{lample2018word}, IS \citep{smith2017offline}, and GC \citep{dinu2015hubness} and three assignment-based methods: Sinkhorn \citep{cuturi2013sinkhorn}, Hungarian matching \citep{kuhn1955hungarian}, and Partial OT \citep{chapel2020partial}. For each representation, all inference methods operate on the same frozen similarity matrix, with independent retrieval serving as the context-invariant reference. The fixed configurations and test-time complexity of all inference methods are provided in Appendix~\ref{app:inference_implementation}.

\textbf{CMRE Evaluation.}
We apply CMRE following the protocol in Section~\ref{sec:cmre}, using the same
constructed contexts for all inference methods. The conventional full-context
setting ($m=200$) is evaluated separately. We report CM-Top1 and CM-Top5 for
all methods. SE is measured by the corresponding
Top-$K$ gain over independent retrieval, whereas CC is
reported only for reduced contexts with $R_m>1$. We use $T=10$ construction
rounds in the main experiments. Analysis up to $T=50$ shows that the estimates
at $T=10$ are close to those obtained with more rounds (Appendix~\ref{app:sampling_convergence}).

\subsection{Full-Context Structural Exploitability}
\label{subsec:full_context}

We first evaluate structural exploitability at the conventional full-context operating point ($m=N_q=200$), where all test queries are processed jointly. For each frozen representation, every inference rule receives the same full query--gallery similarity matrix. The gain over independent retrieval therefore measures how much additional retrieval performance each rule obtains by exploiting cross-query structure at test time.

Table~\ref{tab:full_context_top1} reports full-context Top-1 accuracy for seven inference strategies across six frozen EEG-to-image representations. All six structure-aware inference methods improve accuracy over independent retrieval for every representation, although the gains vary substantially
across representations. The largest gain is obtained by Hungarian matching on HyFI, which increases Top-1 accuracy from 13.70\% to 34.10\%, an improvement of 20.40 percentage points (pp). Averaged across representations, Hungarian achieves the highest Mean SE (+12.96 pp), followed by Sinkhorn (+10.71 pp) and Partial OT (+8.68 pp). The preferred inference rule is representation-dependent. Among the set-level methods, IS performs best on five representations, whereas GC performs best on ATM. Among the assignment-based methods, Hungarian performs best on UBP, BLUR, and HyFI; Sinkhorn performs best on NICE and BrainHIVE; and Partial OT is marginally best on ATM. Representation comparisons are also inference-dependent. Independent retrieval ranks ATM above UBP, whereas every structure-aware inference rule ranks
UBP above ATM. Thus, even at full context, the preferred inference rule depends on the representation, while the apparent representation ordering depends on the inference rule. These ranking dependencies are even more pronounced in the corresponding
full-context Top-5 results (Appendix~\ref{app:full_context_top5}).

These results establish that structural exploitability depends jointly on the frozen representation and the inference rule. To characterize the source of these gains, Appendix~\ref{app:full_context_gain_origin} analyzes Top-1
wrong-to-correct and correct-to-wrong transitions, together with the initial ground-truth ranks of recovered queries. However, full-context results cannot determine whether either the observed gains or comparative rankings remain stable as the available query context changes. We therefore next evaluate structure-aware inference under reduced query contexts using CMRE.


\begin{table*}[t]
\caption{
Full-context Top-1 retrieval accuracy (\%) across six representations. Mean SE denotes the average gain (pp) over independent retrieval.
}
\label{tab:full_context_top1}
\begin{center}
\setlength{\tabcolsep}{5.2pt}
\renewcommand{\arraystretch}{1.00}
\begin{tabular}{llccccccc}
\toprule
Type & Method
& NICE
& ATM
& UBP
& BLUR
& HyFI
& BrainHIVE
& Mean SE\\
\midrule

Independent
& Independent
& 6.95 & 11.50 & 10.45 & 20.05 & 13.70 & 22.25 & -- \\
\midrule

\multirow{3}{*}{Set-level}
& CSLS
& 7.65 & 14.50 & 17.20 & 26.95 & 20.90 & 30.35 & 5.44 \\

& IS
& 8.15 & 14.25 & 21.65 & 30.20 & 27.70 & 35.50 & 8.76 \\

& GC
& 7.10 & 15.60 & 20.95 & 27.20 & 26.05 & 30.90 & 7.15 \\
\midrule

\multirow{3}{*}{Assignment}

& Sinkhorn
& 9.05 & 15.50 & 20.15 & 35.25 & 26.90 & 42.30 & 10.71 \\

& Hungarian
& 7.85 & 15.35 & 23.60 & 40.15 & 34.10 & 41.60 & 12.96 \\

& Partial OT
& 8.20 & 15.55 & 18.75 & 30.90 & 23.40 &40.20 & 8.68 \\
\bottomrule
\end{tabular}
\end{center}
\end{table*}

\subsection{Structural Exploitability under Reduced Query Contexts}
\label{subsec:cm_performance_exp}

The full-context results quantify structural exploitability at a single operating point, but do not show whether retrieval performance, prediction stability, and comparative rankings remain stable as query context changes. We therefore apply CMRE at reduced query-context sizes $m\in\{25,50,100,150\}$. Across these evaluations, the image gallery and frozen full query--gallery similarity matrix remain fixed, while only the size and composition of the constructed query contexts vary.

\begin{figure*}[t]
    \centering
    \includegraphics[width=\textwidth,height=5cm]{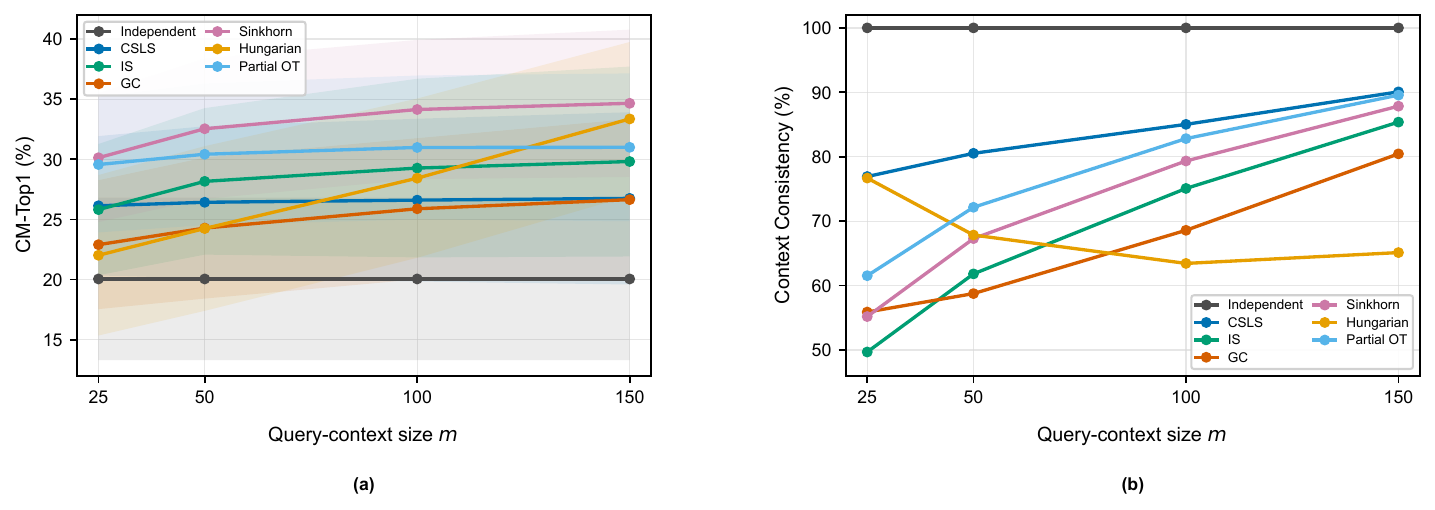}
    \caption{
Context-marginalized retrieval performance and context robustness on BLUR across query-context sizes
$m$.
(a) Context-marginalized Top-1 retrieval accuracy (CM-Top1); shaded
bands indicate standard deviation across subjects.
(b) Same-query Context Consistency (CC)
}
    \label{fig:cm_performance_consistency}
\end{figure*}

Table~\ref{tab:full_context_top1} identifies Hungarian as the full-context
winner on BLUR, achieving 40.15\% Top-1 accuracy at $m=200$. However,
Fig.~\ref{fig:cm_performance_consistency}(a) shows that Sinkhorn ranks first at every reduced
context size, whereas Hungarian declines from 33.36\% at $m=150$ to 22.02\%
at $m=25$. Retaining the full-context winner therefore incurs losses of
1.30, 5.71, 8.29, and 8.11 pp at $m=150$, 100, 50, and 25, respectively.
Independent retrieval remains context-invariant at 20.05\%. Thus, on BLUR,
reducing the query context reverses the preference between Hungarian and
Sinkhorn.

Table~\ref{tab:cm_gain_backbones} summarizes mean CM-Top1 structural exploitability across the four reduced context sizes. All 36 representation--rule combinations yield positive mean gains over independent retrieval, ranging from 0.02 pp for GC on NICE to 12.82 pp for Sinkhorn on BLUR. Thus, useful
cross-query structure remains exploitable even when inference operates on only a subset of the test queries.
To assess whether the observed gains were consistent across subjects, we averaged each subject’s gain across the six representations and four reduced context sizes. All six structure-aware inference rules yielded positive mean subject-level gains, and the corresponding subject-level gains differed significantly from zero under two-sided Wilcoxon signed-rank tests with Holm correction across methods (Appendix~\ref{app:statistical_significance}). However, both the magnitude of the gain and the ranking of inference rules
depend strongly on the frozen representation. Sinkhorn performs best on NICE (+1.32 pp), BLUR (+12.82 pp), and BrainHIVE (+10.02 pp), whereas CSLS ranks first on ATM (+2.51 pp) and IS ranks first on UBP (+10.34 pp) and HyFI (+12.69 pp). No single inference rule is therefore uniformly optimal across representations. The same conclusion extends beyond Top-1 retrieval: CM-Top5 also shows positive structural gains across frozen representations, while the preferred inference rule remains representation-dependent (Appendix~\ref{app:Context_Marginalized_context_top5}).

Averaged across the six representations, Sinkhorn achieves the largest mean reduced-context gain (+7.99 pp), followed by Partial OT (+6.85 pp) and IS (+6.40 pp). Across the 24 representation--context combinations, however, the
full-context winner is no longer optimal in 17 cases (70.8\%). Selecting the full-context winner incurs an average Top-1 loss of 2.70 pp relative to the best method at each reduced context (Appendix~\ref{app:detailed_reduced_context}).
Thus, changes in query context can alter not only retrieval performance but also the ranking of inference rules. We next examine whether same-query predictions remain consistent across contexts.

\begin{table*}[t]
\caption{
CM-Top1 structural exploitability (SE, pp), averaged
over the reduced query-context sizes. The final column
reports the average across the six frozen representations.
}
\label{tab:cm_gain_backbones}
\begin{center}
\renewcommand{\arraystretch}{0.90}

\begin{tabular*}{\textwidth}{
@{\extracolsep{\fill}}
llccccccc
@{}}
\toprule
Type & Method
& NICE
& ATM
& UBP
& BLUR
& HyFI
& BrainHIVE
& Average \\
\midrule

\multirow{3}{*}{Set-level}
& CSLS
& 0.68 & \textbf{2.51} & 6.34 & 6.43 & 7.45 & 7.46 & 5.15 \\

& IS
& 0.36 & 1.14 & \textbf{10.34} & 8.22 & \textbf{12.69} & 5.62 & 6.40 \\

& GC
& 0.02 & 2.28 & 7.37 & 4.88 & 7.91 & 5.22 & 4.61 \\
\midrule

\multirow{3}{*}{Assignment}
& Sinkhorn
& \textbf{1.32} & 2.28 & \underline{9.38} & \textbf{12.82} & \underline{12.14} & \textbf{10.02} & \textbf{7.99} \\

& Hungarian
& 0.46 & 2.36 & 6.78 & 6.96 & 8.37 & 7.01 & 5.32 \\

& Partial OT
& \underline{0.99} & \underline{2.40} & 8.08 & \underline{10.44} & 9.75 & \underline{9.47} & \underline{6.85} \\
\bottomrule
\end{tabular*}
\end{center}
\end{table*}

\subsection{Context Robustness under Reduced Query Contexts}
\label{subsec:context_consistency_exp}
\begin{figure*}[t]
    \centering
    \includegraphics[width=\textwidth,height=5cm]{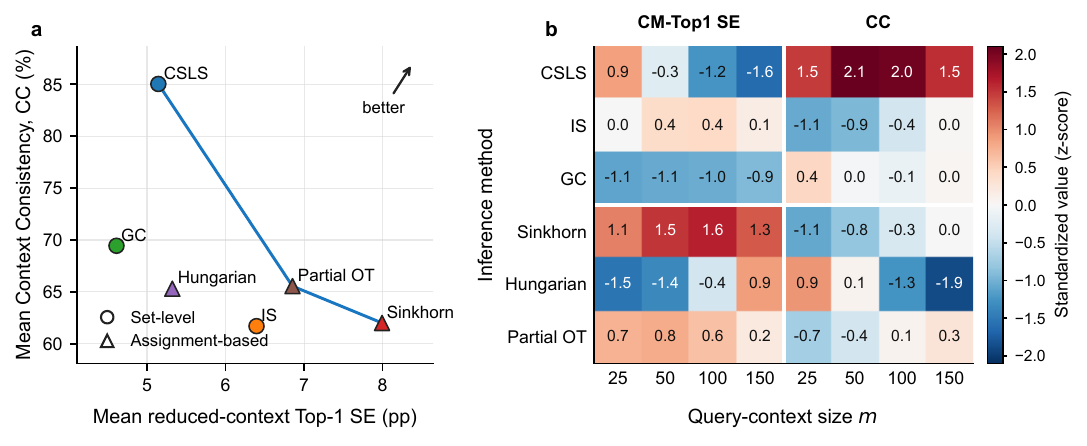}
   \caption{
Joint profiles of structural exploitability and context robustness.
(a) Mean reduced-context Top-1 SE versus mean Context Consistency (CC), across six frozen representations and four reduced context sizes.
(b) Context-size-specific SE and CC, averaged across
representations and standardized across rules within each metric and context size.
}
    \label{fig:gain_robustness_joint}
\end{figure*}
Retrieval performance alone does not reveal whether the same EEG query retains its Top-1 prediction as the companion-query composition changes. We therefore quantify context robustness using Context Consistency
(CC), which measures pairwise agreement among the Top-1 predictions produced for the same query across its constructed context evaluations at a given context size.

Figure~\ref{fig:cm_performance_consistency}(b) reports the CC profiles on
BLUR across the four reduced context sizes. Independent retrieval maintains
$\mathrm{CC}=100\%$ because its predictions are context-invariant by
construction. Among the structure-aware inference rules, CSLS is comparatively
stable, with CC increasing from 76.92\% at $m=25$ to 90.07\% at $m=150$.
By contrast, Hungarian's CM-Top1 increases from 22.02\% to 33.36\% over the
same range, while its CC decreases from 76.69\% to 65.12\%, reaching a minimum
of 63.44\% at $m=100$. This divergence shows that higher retrieval performance
does not necessarily imply greater context robustness.

Table~\ref{tab:cc_backbones} shows that the pattern extends beyond
BLUR. CSLS achieves the highest mean CC on each of the six frozen
representations and the highest overall CC (85.04\%). The other inference
rules attain lower overall CC values, ranging from 61.70\% to 69.44\%, and
vary more strongly across representations. To assess structural exploitability and context robustness jointly,
Fig.~\ref{fig:gain_robustness_joint}(a) plots each inference rule's mean
reduced-context Top-1 gain against its mean CC. CSLS achieves the highest CC
(85.04\%) with a moderate gain (+5.15 pp), whereas Sinkhorn provides the
largest gain (+7.99 pp) but substantially lower CC (62.00\%). Partial OT
occupies an intermediate operating point, combining a relatively large gain
(+6.85 pp) with higher CC (65.56\%) than Sinkhorn. These three methods form
the non-dominated Pareto frontier; IS, GC, and Hungarian are dominated by
methods with both higher gain and higher CC.

Figure~\ref{fig:gain_robustness_joint}(b) resolves these joint profiles by context
size. For each $m$, structural exploitability and CC are averaged across the
six representations and standardized across inference rules within each
metric. Sinkhorn remains above average in structural exploitability but below
average in CC, whereas CSLS exhibits the opposite profile. Hungarian's
relative gain improves as the context size increases, while its relative CC
declines. Thus, no inference rule dominates both dimensions, and method
selection depends on the relative priority assigned to retrieval gain and
context robustness.

Detailed CC results for all six representations and individual context sizes are provided in Appendix~\ref{app:detailed_cc}. Because CC measures agreement irrespective of correctness, Appendix~\ref{app:cc_decomposition} further
decomposes same-query behavior into Stable-Correct, Stable-Wrong, and Context-Sensitive cases. Averaged across representations and context sizes, CSLS exhibits both the highest Stable-Correct fraction (14.90\%) and the
largest Stable-Wrong fraction (48.14\%). High context consistency can therefore reflect both beneficial and erroneous stability and should be interpreted jointly with retrieval performance. 
Beyond query-context variation, we also evaluated robustness to gallery expansion by adding unmatched distractors. On BLUR, structure-aware inference remained beneficial as the gallery grew although the gains diminished (Appendix~\ref{app:gallery_expansion}).













\begin{table*}[t]
\caption{
Average same-query context consistency (CC,\%) across reduced query contexts. Final column reports the average across the six frozen representations. 
}
\label{tab:cc_backbones}
\begin{center}
\small
\renewcommand{\arraystretch}{1.0}
\begin{tabular*}{\textwidth}{@{\extracolsep{\fill}}llccccccc@{}}
\toprule
Type & Method
& NICE
& ATM
& UBP
& BLUR
& HyFI
& BrainHIVE
& Average \\
\midrule

\multirow{3}{*}{Set-level}
& CSLS
& \textbf{83.81} & \textbf{84.81} & \textbf{85.79} & \textbf{83.14} & \textbf{85.78} & \textbf{86.89} & \textbf{85.04} \\

& IS
& 54.78 & 47.32 & 76.23 & 67.99 & 74.72 & 49.17 & 61.70 \\

& GC
& 65.28 & \underline{77.24} & 63.07 & 65.92 & 66.34 & \underline{78.81} & \underline{69.44} \\
\midrule

\multirow{3}{*}{Assignment}
& Sinkhorn
& 58.06 & 43.34 & 77.32 & 72.42 & 75.64 & 45.24 & 62.00 \\

& Hungarian
& \underline{65.69} & 64.66 & 62.03 & 68.27 & 64.01 & 67.17 & 65.31 \\

& Partial OT
& 60.90 & 46.28 & \underline{81.04} & \underline{76.51} & \underline{80.19} & 48.45 & 65.56 \\

\bottomrule
\end{tabular*}
\end{center}
\end{table*}

\subsection{Validation on THINGS-MEG}
\label{subsec:gain_robustness}

\begin{figure}[t]
    \centering
    \includegraphics[width=\linewidth]{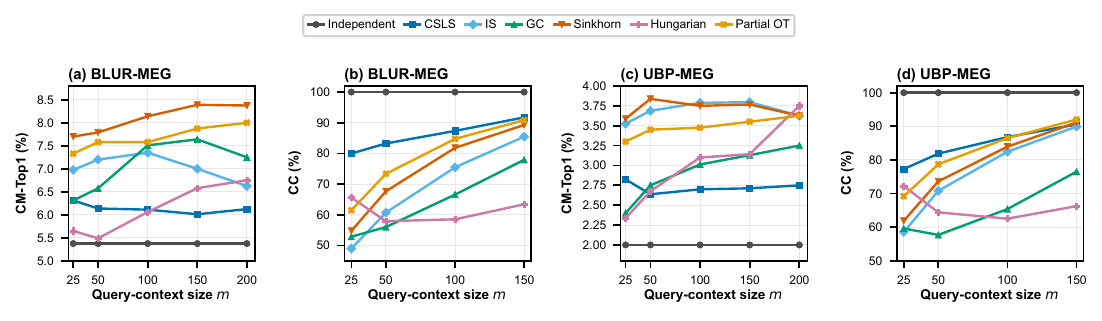}
    \caption{
 CM-Top1 and CC across query-context sizes for BLUR (a,b) and UBP (c,d).
    }
    \label{fig:meg_cmre}
\end{figure}

\begin{table}[t]
\centering
\renewcommand{\arraystretch}{0.80}
\caption{Validation on THINGS-MEG. Full-context Top-1 is averaged across four held-out subjects. Reduced-context results are
averaged over $m\in\{25,50,100,150\}$. $\mathrm{SE}$ denotes Top-1 structural exploitability relative to independent retrieval.
}
\label{tab:meg_summary}
\begin{tabular}{llccccc}
\toprule
\multirow{2}{*}{Representation} &
\multirow{2}{*}{Method} &
\multicolumn{2}{c}{Full Context} &
\multicolumn{3}{c}{Reduced Contexts} \\
\cmidrule(lr){3-4}
\cmidrule(lr){5-7}
& &
Top-1 (\%) & SE (pp) &
CM-Top1 (\%) & SE (pp) & CC (\%) \\
\midrule

\multirow{7}{*}{BLUR}
& Independent & 5.38 & --    & -- & --    & 100.00 \\
& CSLS        & 6.13 & +0.75 & 6.14 & +0.76 & \textbf{85.58} \\
& IS          & 6.63 & +1.25 & 7.13 & +1.75 & 67.67 \\
& GC          & 7.25 & +1.88 & 7.01 & +1.63 & 63.35 \\
& Sinkhorn    & \textbf{8.38} & \textbf{+3.00}
              & \textbf{8.01} & \textbf{+2.63} & 73.39 \\
& Hungarian   & 6.75 & +1.38 & 5.95 & +0.57 & 61.38 \\
& Partial OT  & 8.00 & +2.63 & 7.59 & +2.21 & 77.62 \\
\midrule

\multirow{7}{*}{UBP}
& Independent & 2.00 & --    & -- & --    & 100.00 \\
& CSLS        & 2.75 & +0.75 & 2.72 & +0.72 & \textbf{84.15} \\
& IS          & 3.63 & +1.63 & 3.70 & +1.70 & 75.43 \\
& GC          & 3.25 & +1.25 & 2.82 & +0.82 & 64.80 \\
& Sinkhorn    & 3.63 & +1.63
              & \textbf{3.74} & \textbf{+1.74} & 77.63 \\
& Hungarian   & \textbf{3.75} & \textbf{+1.75}
              & 2.82 & +0.82 & 66.35 \\
& Partial OT  & 3.63 & +1.63 & 3.45 & +1.45 & 81.60 \\
\bottomrule
\end{tabular}
\end{table}

Table~\ref{tab:meg_summary} and Figure~\ref{fig:meg_cmre} extend the CMRE analysis to THINGS-MEG. All six structure-aware inference rules outperform independent retrieval under both full and reduced query contexts, indicating that cross-query structure remains exploitable beyond EEG. On BLUR, Sinkhorn performs best in both the full-context setting (8.38\%, +3.00 pp) and the reduced query contexts (8.01\%, +2.63 pp). On UBP, the preferred rule changes with context: Hungarian performs best under full context (3.75\%,+1.75 pp), whereas Sinkhorn achieves the highest mean CM-Top1 under reduced query contexts (3.74\%, +1.74 pp). Figure~\ref{fig:meg_cmre} further shows that context-response profiles vary across inference rules. For most methods, CC increases with query-context size, but the ranking by CC differs from that by SE. CSLS achieves the highest mean CC on BLUR (85.58\%) and UBP (84.15\%), despite modest SE values of +0.76 and +0.72 pp, respectively. These results show that the separation between structural exploitability and context robustness persists in MEG, and that the preferred inference rule depends on both query-context size and representation.

\section{Conclusion}

We studied how test-time structure-aware inference in EEG-to-image retrieval depends on query-context composition. Using fixed EEG–image similarity matrices, we introduced CMRE to measure structural exploitability and Context Consistency across sampled contexts. The preferred inference rule varied across representations and context sizes, while representation rankings were themselves inference- and context-dependent.  Larger structural gains did not necessarily imply greater context robustness, and this separation also persisted on THINGS-MEG. These findings show that structural exploitability and context robustness are complementary properties and that full-context accuracy alone can provide an incomplete evaluation of context-dependent test-time inference. Accordingly, CMRE evaluates inference across sampled query contexts by jointly considering context-marginalized retrieval performance, structural gains, and same-query prediction consistency.

\subsection*{AI use statement}
In this work, generative AI tools were used to assist with language editing, LaTeX formatting, visualization refinement, and limited code development. All AI-assisted outputs were reviewed and validated by the authors. The authors independently verified all analyses, results, figures, and scientific claims and take full responsibility for the final manuscript.

\subsection*{Reproducibility statement}

All experiments use a fixed evaluation pipeline in which each inference
rule operates on frozen EEG-to-image query--gallery similarity matrices,
without updating either encoder. Section~\ref{sec:cmre} and Appendix~\ref{app:cmre_algorithm} specify the CMRE formulation, balanced context-construction schedule, inference rules, and evaluation metrics.


\bibliography{iclr2027_conference}
\bibliographystyle{iclr2027_conference}

\appendix
\section{Related Work}
\label{sec:related_work}
\subsection{EEG-to-Image Retrieval}

Most EEG-to-image retrieval methods align EEG embeddings with pretrained visual spaces. Early studies established zero-shot retrieval through
contrastive neural--visual learning~\citep{song2024decoding,li2024visual,
wei2024mb2c,chen2024muse}. Subsequent work added language-guided supervision,
multimodal semantic bridging, and richer cross-modal
objectives~\citep{song2025nicepp,li2025neuralmcrl,
zhang2025cognitioncapturer,chen2026mindalign,wang2026cortiva}. More recent methods explored prompt tuning, hierarchical visual representations,
pretrained vision encoders, and subject-aware
alignment~\citep{wang2025neuroclip,zhou2026eegit,lee2026definealignfuse,
jiang2026samga}.

Perceptual and neuroscientific priors have also been introduced through
uncertainty-aware supervision, multiscale targets, hyperbolic modeling,
neuromimetic simulation, and masked cross-modal
fusion~\citep{wu2025ubp,zheng2026brainhive,jo2026hyfi,
zhang2026neurobridge,shao2026bicap,xu2026neurovista}. Cross-subject retrieval remains challenging because EEG representations vary
across individuals. Recent methods address this problem through
subject-invariant learning, label-free calibration, similarity-geometry
adaptation, and target-subject coordinate
recovery~\citep{choi2026subjectinvariant,huang2026sattc,cui2026score}.
A recent diagnostic study further showed that standard retrieval performance
may not transfer to controlled evaluation
settings~\citep{zhang2026eegeditbench}. However, prior work generally
evaluates each target under a fixed test context, leaving sensitivity to
companion-query composition insufficiently characterized.

\subsection{Test-Time Structural Inference for Retrieval}

Post-hoc structural inference refines retrieval scores without changing the
underlying representation. One line of work addresses hubness, where a
small number of gallery items become nearest neighbors of many
queries~\citep{dinu2015hubness,smith2017offline,lample2018word,
bogolin2022qbnorm,wang2023dbnorm,lin2025neighborretr,pan2025dbsn}.
Global Correction (GC) re-ranks candidates using query-population
preferences~\citep{dinu2015hubness}; Inverted Softmax (IS) suppresses popular
targets through reverse normalization~\citep{smith2017offline}; and
Cross-domain Similarity Local Scaling (CSLS) corrects for local embedding
density~\citep{lample2018word}. Related matrix-normalization methods have also
been applied to cross-modal retrieval~\citep{cheng2021dualsoftmax,
bogolin2022qbnorm,wang2023dbnorm}. We refer to these methods as
set-level inference because predictions remain query-wise while using
statistics from other queries or gallery items.

Joint matching introduces stronger structural coupling. Optimal transport
provides soft global assignments~\citep{cuturi2013sinkhorn,courty2017optimal,
chizat2018unbalanced,chapel2020partial}, with entropic regularization enabling
efficient Sinkhorn iterations~\citep{cuturi2013sinkhorn}. Unbalanced transport
relaxes mass conservation~\citep{chizat2018unbalanced}, while partial
transport relaxes full matching~\citep{chapel2020partial}; the Hungarian
algorithm instead imposes hard one-to-one
assignment~\citep{kuhn1955hungarian}. Related global matching methods have
also been used in vision and cross-modal
correspondence~\citep{sarlin2020superglue,cheng2022cgm}. These approaches span
score recalibration, soft global coupling, and hard assignment. The effectiveness of post-hoc normalization can also depend on available
query and gallery statistics. Querybank Normalisation uses an external query
bank~\citep{bogolin2022qbnorm}, whereas Dual Bank Normalization uses both query
and gallery banks~\citep{wang2023dbnorm}. Target-set structure has also been
used for label-free cross-subject EEG-to-image
calibration~\citep{huang2026sattc,cui2026score}. However, these studies
optimize retrieval under prescribed test configurations. We instead hold the
representation and inference rule fixed and vary only the available query
context.

\subsection{Evaluation of Transductive Inference}

Transductive inference uses statistics from unlabeled test samples, making its
performance sensitive to assumptions about the query set. Transductive
few-shot methods benefit from joint reasoning over
queries~\citep{boudiaf2020tim}, but evaluations show that class balance,
class composition, and query-set structure can substantially alter measured
performance~\citep{veilleux2021realistic,martin2022practical}. Test-set
composition therefore forms part of the information available to an inference
procedure rather than merely a benchmark detail.

Related concerns arise in test-time adaptation. Early methods adapt models
using statistics or entropy objectives from unlabeled
batches~\citep{wang2021tent}, while continual and source-free variants address
changing test distributions~\citep{wang2022cotta,niu2022eata}. Test-time
behavior is also affected by temporal correlation,
batch size, class imbalance, and distribution
shifts~\citep{gong2022note,niu2023sar,yuan2023robust}. Benchmarks further show
sensitivity to evaluation protocols, model selection, backbone quality, and
shift type~\citep{zhao2023pitfalls,wang2025otta,hagen2024queryvariation,
sun2026benchcir}. However, these studies primarily vary distributions or
adaptation settings rather than repeatedly evaluating the same target under
controlled changes to its companion-query context.

\section{Implementation Details}
\label{app:implementation_details}

This section provides additional implementation details for the six
EEG-to-image retrieval representations and the six test-time structural inference
methods used in our experiments. For each representation, we preserve its original
EEG--image representation pipeline and obtain a frozen query--gallery
similarity matrix before applying any structural inference. The inference
methods therefore modify only the test-time similarity structure and do not
update the representation parameters or embeddings.

\subsection{Representation Implementations}
\label{app:backbone_implementations}

\paragraph{NICE.}
NICE~\citep{song2024decoding} represents the standard contrastive
EEG--vision alignment paradigm. It learns an EEG encoder whose outputs are
directly aligned with pretrained visual representations through a contrastive
objective, providing a relatively direct neural-to-visual embedding space.

\paragraph{ATM.}
ATM~\citep{li2024visual} introduces an Adaptive Thinking Mapper to transform
EEG representations toward the pretrained visual embedding space. Compared
with the direct contrastive alignment of NICE, ATM explicitly learns a
mapping between the neural and visual representations, providing a different
similarity geometry for retrieval. We use the retrieval branch of the
original framework without involving its reconstruction components.

\paragraph{UBP.}
UBP~\citep{wu2025ubp} incorporates uncertainty-aware visual supervision into
EEG--image alignment. Rather than treating every EEG--image pair as an equally
reliable training target, it modulates visual supervision according to the
estimated ambiguity of neural--visual correspondence. This produces a
similarity space shaped by uncertainty-aware alignment rather than purely
deterministic contrastive matching.

\paragraph{BLUR.}
BLUR~\citep{liu2026leveraging} constructs a perceptually diversified visual
space using multiple levels of image blur and learns to select informative
visual evidence for EEG alignment. Its retrieval representation therefore
reflects multi-view perceptual supervision rather than a single visual target,
making it particularly useful for examining whether test-time structural
inference behaves differently on a perceptually enriched similarity space.

\paragraph{HyFI.}
HyFI~\citep{jo2026hyfi} models EEG--visual relations in a hyperbolic feature
space, aiming to capture hierarchical semantic structure that may be difficult
to represent in a purely Euclidean embedding. It therefore provides a
geometrically distinct retrieval representation from the Euclidean
contrastive and perceptual-alignment backbones considered above.

\paragraph{BrainHIVE.}
BrainHIVE~\citep{zheng2026brainhive} employs hierarchical and multiscale
visual supervision to align EEG representations with visual information at
multiple semantic granularities. Unlike approaches based on a single visual
embedding or a single perceptual transformation, BrainHIVE integrates
complementary visual representations before retrieval, yielding a richer
hierarchical EEG--image similarity structure.

The six representations are selected to span different representation families
rather than merely different model capacities. NICE and ATM provide relatively
direct neural--visual alignment, UBP and BLUR modify visual supervision through
perceptual degradation, HyFI introduces a different embedding geometry, and
BrainHIVE incorporates hierarchical multi-source visual representations.
This diversity allows us to evaluate whether structural inference behavior
persists across different frozen EEG--image similarity geometries.

\subsection{Test-Time Inference Methods}
\label{app:inference_implementations}
All inference methods operate on the same frozen EEG--image similarity matrix
$\mathbf{S}\in\mathbb{R}^{N_q\times N_g}$, where
$S_{ij}$ denotes the cosine similarity between EEG query $q_i$ and gallery
candidate $g_j$. We group the compared approaches into set-level methods, which recalibrate retrieval scores using statistics from the available
query set, and assignment-based methods, which jointly couple queries
and gallery candidates through global matching constraints.

\paragraph{CSLS.}
Cross-domain Similarity Local Scaling (CSLS)
~\citep{lample2018word} mitigates hubness by recalibrating each pairwise
similarity using local neighborhood statistics on both sides of the retrieval
space. For query $i$ and gallery candidate $j$, the adjusted similarity is
\begin{equation}
S_{ij}^{\mathrm{CSLS}}
=
2S_{ij}-r_q(i)-r_g(j),
\end{equation}
where
\begin{equation}
r_q(i)
=
\frac{1}{k}
\sum_{j'\in\mathcal{N}_g^k(i)} S_{ij'},
\qquad
r_g(j)
=
\frac{1}{k}
\sum_{i'\in\mathcal{N}_q^k(j)} S_{i'j}.
\end{equation}
Here, $\mathcal{N}_g^k(i)$ and $\mathcal{N}_q^k(j)$ denote the corresponding $k$-nearest neighborhoods. Retrieval is then performed using the recalibrated scores $S_{ij}^{\mathrm{CSLS}}$.

\paragraph{Inverted Softmax.}
Inverted Softmax (IS)~\citep{smith2017offline} suppresses gallery candidates
that receive uniformly high similarity from many competing queries. For each
query--gallery pair, we compute
\begin{equation}
S_{ij}^{\mathrm{IS}}
=
\frac{\exp(\beta S_{ij})}
{\sum_{i'\neq i}\exp(\beta S_{i'j})},
\end{equation}
where $\beta$ controls the sharpness of the normalization. The denominator
captures the relative popularity of gallery candidate $j$ across the available
queries, while the target query itself is excluded from the normalization.

\paragraph{Global Correction.}
Global Correction (GC)~\citep{dinu2015hubness} corrects hubness by
reversing the ranking perspective. For gallery candidate $j$, let
\begin{equation}
\rho_{ij}
=
1+
\sum_{i'\neq i}
\mathbb{I}\!\left[S_{i'j}>S_{ij}\right]
\end{equation}
denote the rank of query $q_i$ among all competing queries with respect to
candidate $g_j$. GC predicts
\begin{equation}
\hat{y}_{i}^{\mathrm{GC}}
=
\arg\min_j \rho_{ij},
\end{equation}
with the original similarity $S_{ij}$ used to resolve rank ties. Candidates
that rank the target query unusually highly are therefore preferred over
globally popular candidates.

\paragraph{Sinkhorn.}
Sinkhorn inference~\citep{cuturi2013sinkhorn} performs globally coupled soft
matching through entropy-regularized optimal transport. We define the transport
cost as
\begin{equation}
C_{ij}=-S_{ij},
\end{equation}
and solve
\begin{equation}
\mathbf{P}^{*}
=
\arg\min_{\mathbf{P}\in\Pi(\mathbf{a},\mathbf{b})}
\left\langle\mathbf{P},\mathbf{C}\right\rangle
-
\varepsilon H(\mathbf{P}),
\end{equation}
where
\begin{equation}
\Pi(\mathbf{a},\mathbf{b})
=
\left\{
\mathbf{P}\geq0:
\mathbf{P}\mathbf{1}=\mathbf{a},
\mathbf{P}^{\top}\mathbf{1}=\mathbf{b}
\right\},
\end{equation}
and
\begin{equation}
H(\mathbf{P})
=
-\sum_{i,j}P_{ij}\log P_{ij}.
\end{equation}
We use uniform marginals
$\mathbf{a}=\frac{1}{N_q}\mathbf{1}$ and
$\mathbf{b}=\frac{1}{N_g}\mathbf{1}$.
For a reduced context $B$ with $|B|=m$, the 
marginals is defined as
$\mathbf a_B=\frac{1}{m}\mathbf 1_m$. The transport plan $\mathbf{P}^{*}$ is obtained using Sinkhorn iterations, and
gallery candidates are ranked according to the corresponding row of
$\mathbf{P}^{*}$.

\paragraph{Hungarian Matching.}
Hungarian matching~\citep{kuhn1955hungarian} performs hard global assignment
by maximizing the total similarity between queries and gallery candidates:
\begin{equation}
\mathbf{X}^{*}
=
\arg\max_{\mathbf{X}}
\sum_{i=1}^{N_q}
\sum_{j=1}^{N_g}
X_{ij}S_{ij},
\end{equation}
subject to
\begin{equation}
X_{ij}\in\{0,1\},
\qquad
\sum_{j=1}^{N_g}X_{ij}=1,
\qquad
\sum_{i=1}^{N_q}X_{ij}\leq1.
\end{equation}
Thus, every query is assigned exactly one gallery candidate, while each
candidate can be assigned to at most one query. This formulation also applies
to the expanded-gallery setting where $N_g>N_q$.

\paragraph{Partial Optimal Transport.}
Partial Optimal Transport (Partial OT)~\citep{chapel2020partial} relaxes the
full mass-conservation assumption of standard optimal transport and transports
only a prescribed fraction of the available mass. Using the same cost matrix
$\mathbf{C}=-\mathbf{S}$, we solve
\begin{equation}
\mathbf{P}^{*}
=
\arg\min_{\mathbf{P}\geq0}
\left\langle\mathbf{P},\mathbf{C}\right\rangle
-
\varepsilon H(\mathbf{P}),
\end{equation}
subject to
\begin{equation}
\mathbf{P}\mathbf{1}\leq\mathbf{a},
\qquad
\mathbf{P}^{\top}\mathbf{1}\leq\mathbf{b},
\qquad
\mathbf{1}^{\top}\mathbf{P}\mathbf{1}=\tau.
\end{equation}
Here, $\tau$ controls the amount of transported mass. Unlike full transport,
Partial OT does not require all query and gallery mass to participate in the
coupling, making it more flexible in the presence of unmatched candidates.

\subsection{Test-Time Inference Configurations and Complexity}
\label{app:inference_implementation}
To enable a controlled comparison, we fix each inference rule and its associated hyperparameters across all frozen representations and evaluation settings, including all context sizes. This design isolates the effect of available query context from context-specific retuning. Accordingly, our results characterize the behavior of each rule at a fixed operating point, rather than its optimally tuned performance for every representation or context size. Table~\ref{tab:inference_config_complexity} summarizes the corresponding configurations and asymptotic inference complexity after the $N_q\times N_g$ similarity matrix has been constructed.

\begin{table}[t]
\centering
\caption{
Fixed configurations and test-time complexity of the compared inference
methods. $N_q$ and $N_g$ denote the numbers of EEG queries and gallery
candidates, respectively, and $L$ denotes the number of transport iterations.
}
\label{tab:inference_config_complexity}

\renewcommand{\arraystretch}{1.08}
\begin{tabular}{@{}l l c@{}}
\toprule
Method & Fixed configuration & Complexity \\
\midrule
Independent & None                            & $O(N_qN_g)$ \\
CSLS        & $k=10$                          & $O(N_qN_g)$ \\
IS          & $\beta=30$                      & $O(N_qN_g)$ \\
GC          & None                            & $O(N_qN_g\log N_q)$ \\
Sinkhorn    & $\varepsilon=0.05$              & $O(LN_qN_g)$ \\
Partial OT  & $\tau=0.90,\ \varepsilon=0.05$ & $O(LN_qN_g)$ \\
Hungarian   & None                            & $O(N_q^2N_g)$ \\
\bottomrule
\end{tabular}
\end{table}
Independent retrieval, CSLS, and IS have costs linear in the similarity-matrix size. GC additionally performs candidate-to-query sorting, resulting in $O(N_qN_g\log N_q)$ complexity. Sinkhorn and Partial OT require iterative global coupling with complexity $O(LN_qN_g)$, where $L$ is the number of transport iterations. Transport solvers use $L_{\max}=10{,}000$ and a convergence tolerance of $10^{-8}$. Hungarian matching performs hard global assignment with complexity $O(N_q^2N_g)$ for $N_q\leq N_g$. Importantly, these computations are performed only at test time on already extracted embeddings. No method requires retraining or fine-tuning the EEG or image encoders. Using the same fixed configurations across all representations also prevents specific hyperparameter tuning from confounding the comparison of inference behavior.

\subsection{CMRE Evaluation Procedure}
\label{app:cmre_algorithm}

Algorithm~\ref{alg:cmre} summarizes the complete CMRE evaluation
procedure, including context construction, structure-aware inference,
and metric aggregation. Table~\ref{tab:cmre_T_Rm} further specifies
the context-construction schedule used in our experiments and the
resulting relationship between $T$ and $R_m$.

\begin{algorithm}[t]
\caption{Context-Marginalized Retrieval Evaluation (CMRE)}
\label{alg:cmre}

\begin{algorithmic}

\State \textbf{Input:}
Frozen similarity matrix
$\mathbf{S}\in\mathbb{R}^{N_q\times N_g}$,
ground-truth targets $\mathbf{y}$,
inference rule $\Phi$, context size $m$,
rank cutoff $K$, and construction rounds $T$.

\State \textbf{Output:}
$\mathrm{CM\text{-}TopK}_{\Phi}(m)$ and,
when defined, $\mathrm{CC}_{\Phi}(m)$.

\Statex
\Statex \textbf{// Query-balanced context construction}

\State Initialize $\mathcal{R}_i\gets[\,]$ and
$\mathcal{P}_i\gets[\,]$ for all $i=1,\ldots,N_q$.

\If{$m=N_q$}
    \State $T\gets 1$
\EndIf

\For{$t=1,\ldots,T$}
    \State Construct the query-balanced context collection
    $\mathcal{C}_t^{(m)}$.

    \For{each context $\mathcal{B}\in\mathcal{C}_t^{(m)}$}
        \State $\mathbf{S}_{\mathcal{B}}
        \gets \mathbf{S}[\mathcal{B},:]$

        \State Obtain query-wise gallery rankings
        $\{\boldsymbol{\pi}_{i,\mathcal{B}}\}_{q_i\in\mathcal{B}}$
        from $\Phi(\mathbf{S}_{\mathcal{B}})$.

        \For{each query $q_i\in\mathcal{B}$}
            \State $\rho_{i,\mathcal{B}}
            \gets
            \operatorname{rank}_{\boldsymbol{\pi}_{i,\mathcal{B}}}(y_i)$

            \State $\hat{y}_{i,\mathcal{B}}
            \gets \operatorname{top1}(\boldsymbol{\pi}_{i,\mathcal{B}})$

            \State Append $\rho_{i,\mathcal{B}}$ to $\mathcal{R}_i$.
            \State Append $\hat{y}_{i,\mathcal{B}}$ to $\mathcal{P}_i$.
        \EndFor
    \EndFor
\EndFor

\Statex
\Statex \textbf{// Metric aggregation}

\State Compute $\mathrm{CM\text{-}TopK}_{\Phi}(m)$ from
$\{\mathcal{R}_i\}_{i=1}^{N_q}$.

\State $R_m\gets|\mathcal{P}_1|$

\If{$R_m>1$}
    \State Compute $\mathrm{CC}_{\Phi}(m)$ from
    $\{\mathcal{P}_i\}_{i=1}^{N_q}$.
\Else
    \State $\mathrm{CC}_{\Phi}(m)\gets\mathrm{undefined}$.
\EndIf

\State \Return
$\mathrm{CM\text{-}TopK}_{\Phi}(m)$ and
$\mathrm{CC}_{\Phi}(m)$ when defined.

\end{algorithmic}
\end{algorithm}

\begin{table}[t]
\centering
\caption{Relationship between partition rounds and per-query context counts in CMRE.}
\label{tab:cmre_T_Rm}

\begin{tabular}{ccccc}
\toprule
Context & Contexts & Query appearances &
\multirow[c]{2}{*}{$T$} &
\multirow[c]{2}{*}{$R_m$} \\
size $m$ & per round & per round & & \\
\midrule
25  & 8 & 1 & 10 & 10 \\
50  & 4 & 1 & 10 & 10 \\
100 & 2 & 1 & 10 & 10 \\
150 & 4 & 3 & 10 & 30 \\
\bottomrule
\end{tabular}
\end{table}

\section{Additional Full-Context Results}
\label{app:full_context_results}

\subsection{Analysis of Full-Context Structural Gains}
\label{app:full_context_gain_origin}

To characterize the full-context Top-1 gains, we analyze how predictions change relative to independent retrieval. We denote wrong-to-correct and correct-to-wrong transitions by W$\rightarrow$C and C$\rightarrow$W, respectively. For each structural inference method, we further examine the rank of the ground-truth (GT) image under independent retrieval for W$\rightarrow$C cases.

\begin{figure*}[t]
    \centering
    \includegraphics[width=0.98\textwidth]
    {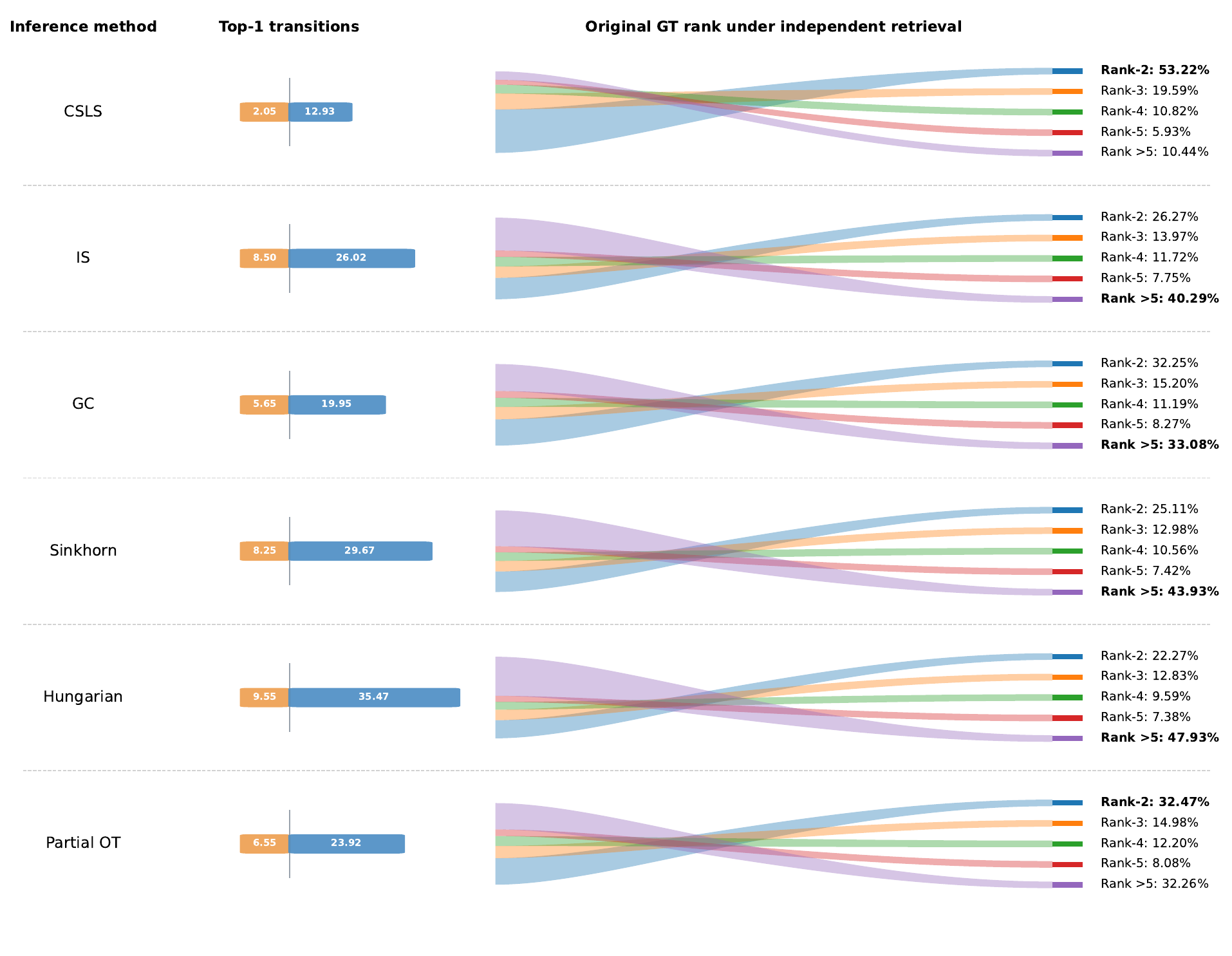}
    \caption{
    Origin of full-context Top-1 structural gains.
   For each inference method, the diverging bars show the average numbers of wrong-to-correct (W$\rightarrow$C; blue) and correct-to-wrong (C$\rightarrow$W; orange) transitions relative to independent retrieval. The flows show the ground-truth (GT) rank under independent retrieval for W$\rightarrow$C queries. Results are averaged across six representations and ten LOSO folds.
    }
    \label{fig:full_context_gain_origin}
\end{figure*}

Figure~\ref{fig:full_context_gain_origin} reveals distinct prediction-transition patterns across inference rules. CSLS primarily corrects near-miss errors: in 89.56\% of its W$\rightarrow$C cases, the GT image was already within the independent Top-5, including 53.22\% at rank 2. By contrast, 43.93\% and 47.93\% of the queries corrected by Sinkhorn and Hungarian, respectively, had initial GT ranks beyond 5. Hungarian produces the most W$\rightarrow$C transitions but also the most C$\rightarrow$W degradations. These patterns show that inference rules exploit the frozen similarity structure differently. Set-level score recalibration mainly resolves near-top ambiguities, whereas globally coupled assignment can recover targets from deeper ranks but may also introduce more errors.

\subsection{Full-Context Top-5 Retrieval Results}
\label{app:full_context_top5}

To complement the full-context Top-1 results reported in the main paper, we evaluate Top-5 retrieval accuracy under the same frozen similarity matrices. Table~\ref{tab:full_context_top5} reports the results for all inference methods that produce a ranked candidate list.

\begin{table*}[t]
\caption{
Full-context Top-5 retrieval accuracy (\%) across six representations. Mean SE denotes the average gain (pp) over independent retrieval. Top-5 is not reported for Hungarian because
hard one-to-one assignment produces a single discrete assignment rather than
a ranked candidate list.
}
\label{tab:full_context_top5}
\begin{center}
\setlength{\tabcolsep}{5.2pt}
\renewcommand{\arraystretch}{1.00}

\begin{tabular}{llccccccc}
\toprule
Type & Method
& NICE
& ATM
& UBP
& BLUR
& HyFI
& BrainHIVE
& Mean SE \\
\midrule

Independent
& Independent
& 20.95
& 30.60
& 30.20
& 49.25
& 36.05
& 47.10
& -- \\
\midrule

\multirow{3}{*}{Set-level}
& CSLS
& 22.95
& 36.35
& 41.80
& 58.65
& 50.75
& 57.85
& 9.03 \\

& IS
& 24.10
& 37.60
& 47.40
& 60.95
& 57.40
& 65.10
& 13.07 \\

& GC
& 24.55
& 40.35
& 49.20
& 58.60
& 59.60
& 64.95
& 13.85 \\
\midrule

\multirow{3}{*}{Assignment}
& Sinkhorn
& 27.90
& 41.30
& 46.50
& 66.90
& 56.65
& 73.15
& 16.38 \\

& Hungarian
& --
& --
& --
& --
& --
& --
& -- \\

& Partial OT
& 26.30
& 41.20
& 42.15
& 61.95
& 52.80
& 69.55
& 13.30 \\
\bottomrule
\end{tabular}
\end{center}
\end{table*}

The Top-5 results show that the benefit of full-context structural inference extends beyond the final Top-1 decision. All structural methods with ranked outputs improve over independent retrieval across all six representations. Averaged across representations, Sinkhorn achieves the largest Top-5 gain (+16.38 pp), followed by GC (+13.85 pp), Partial OT (+13.30 pp), and IS (+13.07 pp). Sinkhorn obtains the highest Top-5 accuracy on NICE, ATM, BLUR, and BrainHIVE, whereas GC performs best on UBP and HyFI.  ATM slightly outperforms UBP (30.60\% vs.\ 30.20\%), whereas UBP surpasses ATM under every structure-aware method that produces a ranked candidate list. Likewise, BLUR outperforms BrainHIVE under independent retrieval, BrainHIVE achieves markedly higher Top-5 accuracy under all structure-aware methods except CSLS. Under GC, HyFI also slightly surpasses BLUR (59.60\% vs.\ 58.60\%). Together, these results show that structural inference affects not only Top-5 retrieval performance but also the apparent ordering of frozen representations, providing further evidence that representation rankings are inference-dependent.

\section{Additional Reduced-Context Results}
\subsection{Detailed Results across Context Sizes}
\label{app:detailed_reduced_context}
Table~\ref{tab:detailed_cm_top1} reports the complete CM-Top1 accuracy at each reduced query-context
size, complementing the context-averaged results in the main paper.
Importantly, the inference rule selected as best under full-context
evaluation does not remain optimal in 17 of the 24
representation--context-size conditions ($70.8\%$). Relative to selecting
the best rule at each reduced context size, using the full-context winner
incurs an average performance loss of $2.70$ pp and a maximum
loss of $8.29$ pp. This instability is observed across
multiple representations: the full-context winner changes in $1/4$ reduced
contexts for NICE, $4/4$ for ATM, $4/4$ for UBP, $4/4$ for BLUR, $3/4$ for
HyFI, and $1/4$ for BrainHIVE. These results show that inference-rule
selection based on a single full-context operating point does not reliably
transfer to reduced-query contexts.

The detailed results further reveal context-response patterns that are
obscured by averaging across context sizes. Some methods are comparatively
insensitive to $m$; for example, CSLS on BLUR changes only from $26.13\%$
at $m=25$ to $26.75\%$ at $m=150$. In contrast, Hungarian on BLUR increases
from $22.02\%$ to $33.36\%$, Hungarian on HyFI from $16.79\%$ to $28.53\%$,
and Sinkhorn on BrainHIVE from $23.27\%$ to $39.33\%$. Thus, similar context-averaged performance can mask substantially different sensitivity to the amount of available query context.

To assess whether the observed ranking reversals reflect negligible accuracy differences, we conduct additional tie-aware analyses. Among the 24 representation--context-size conditions, 16 incur selection regret greater than 0.1 pp and 12 exceed 1 pp. When rules within 0.1 pp of the best are treated as co-optimal in both full and reduced contexts, 13/24 conditions still exhibit disjoint sets of optimal rules. We further perform 10,000 paired subject-level bootstrap resamples, recomputing rule selection and regret across the ten LOSO subjects. The resulting 95\% confidence intervals are [14,19]/24 for ranking reversals and [2.38,3.08] pp for mean selection regret. These analyses indicate that the aggregate selection-reliability findings are not solely driven by near-tied comparisons and remain robust under subject-level resampling.

\begin{table*}[t]
\centering
\caption{
Detailed CM-Top1 accuracy (\%) under reduced query
contexts $m\in\{25,50,100,150\}$. Each entry reports mean$\pm$standard deviation across all ten LOSO folds.
}
\label{tab:detailed_cm_top1}
\begin{center}

\setlength{\tabcolsep}{4.0pt}
\renewcommand{\arraystretch}{1.00}

\begin{tabular*}{\textwidth}{
@{\extracolsep{\fill}}
llcccc
@{}}
\toprule
Representation & Method
& $m=25$
& $m=50$
& $m=100$
& $m=150$ \\
\midrule

\multirow{6}{*}{NICE}
& CSLS
& $\mathbf{7.67\pm2.31}$
& $7.63\pm2.40$
& $7.66\pm2.50$
& $7.58\pm2.63$ \\

& IS
& $6.99\pm1.37$
& $7.29\pm1.69$
& $7.28\pm1.97$
& $7.67\pm2.18$ \\

& GC
& $6.88\pm2.16$
& $7.07\pm2.09$
& $6.91\pm2.00$
& $7.01\pm2.32$ \\

& Sinkhorn
& $\underline{7.61\pm1.76}$
& $\mathbf{8.28\pm1.82}$
& $\mathbf{8.47\pm2.06}$
& $\mathbf{8.71\pm2.18}$ \\

& Hungarian
& $7.07\pm2.24$
& $7.42\pm2.06$
& $7.55\pm2.12$
& $7.61\pm2.38$ \\

& Partial OT
& $7.53\pm1.68$
& $\underline{7.90\pm1.93}$
& $\underline{8.14\pm2.06}$
& $\underline{8.18\pm2.29}$ \\
\midrule

\multirow{6}{*}{ATM}
& CSLS
& $\mathbf{13.61\pm3.16}$
& $\mathbf{13.87\pm3.26}$
& $14.29\pm3.42$
& $14.24\pm3.59$ \\

& IS
& $10.29\pm2.05$
& $12.47\pm2.61$
& $13.74\pm2.94$
& $14.04\pm3.15$ \\

& GC
& $12.33\pm3.53$
& $13.38\pm3.50$
& $14.39\pm3.50$
& $15.01\pm3.62$ \\

& Sinkhorn
& $11.06\pm1.94$
& $13.67\pm2.81$
& $\underline{14.99\pm2.73}$
& $\mathbf{15.40\pm2.95}$ \\

& Hungarian
& $\underline{12.50\pm3.32}$
& $13.29\pm3.24$
& $14.45\pm3.34$
& $15.20\pm3.87$ \\

& Partial OT
& $11.30\pm2.08$
& $\underline{13.82\pm2.71}$
& $\mathbf{15.12\pm3.12}$
& $\underline{15.37\pm3.15}$ \\
\midrule

\multirow{6}{*}{UBP}
& CSLS
& $16.30\pm4.06$
& $16.66\pm4.08$
& $17.01\pm4.45$
& $17.19\pm3.47$ \\

& IS
& $\mathbf{19.79\pm4.07}$
& $\mathbf{20.57\pm4.10}$
& $\mathbf{21.23\pm4.39}$
& $\mathbf{21.59\pm3.59}$ \\

& GC
& $14.99\pm3.80$
& $16.79\pm4.10$
& $19.19\pm5.57$
& $20.29\pm4.92$ \\

& Sinkhorn
& $\underline{19.04\pm3.59}$
& $\underline{19.89\pm3.44}$
& $\underline{20.15\pm3.44}$
& $20.22\pm2.65$ \\

& Hungarian
& $13.95\pm3.19$
& $15.57\pm4.02$
& $18.48\pm5.18$
& $\underline{20.91\pm4.86}$ \\

& Partial OT
& $18.11\pm3.11$
& $18.60\pm3.15$
& $18.69\pm2.90$
& $18.71\pm2.64$ \\
\midrule

\multirow{6}{*}{BLUR}
& CSLS
& $26.13\pm5.79$
& $26.43\pm6.32$
& $26.61\pm6.77$
& $26.75\pm6.80$ \\

& IS
& $25.81\pm5.46$
& $28.16\pm6.09$
& $29.27\pm7.43$
& $29.82\pm7.49$ \\

& GC
& $22.90\pm5.36$
& $24.28\pm5.86$
& $25.89\pm5.87$
& $26.64\pm6.35$ \\

& Sinkhorn
& $\mathbf{30.13\pm5.35}$
& $\mathbf{32.54\pm5.83}$
& $\mathbf{34.14\pm5.79}$
& $\mathbf{34.66\pm5.81}$ \\

& Hungarian
& $22.02\pm6.67$
& $24.25\pm6.85$
& $28.43\pm6.62$
& $\underline{33.36\pm6.07}$ \\

& Partial OT
& $\underline{29.57\pm5.64}$
& $\underline{30.42\pm5.89}$
& $\underline{30.98\pm5.98}$
& $31.00\pm5.83$ \\
\midrule

\multirow{6}{*}{HyFI}
& CSLS
& $20.99\pm4.37$
& $21.21\pm4.94$
& $21.21\pm5.31$
& $21.18\pm5.48$ \\

& IS
& $\mathbf{24.71\pm4.27}$
& $\mathbf{26.24\pm4.40}$
& $\mathbf{27.16\pm4.78}$
& $\underline{27.44\pm5.20}$ \\

& GC
& $17.90\pm4.73$
& $20.65\pm5.04$
& $23.11\pm5.98$
& $24.80\pm6.68$ \\

& Sinkhorn
& $\underline{24.12\pm3.86}$
& $\underline{25.62\pm4.02}$
& $\underline{26.71\pm4.34}$
& $26.90\pm4.58$ \\

& Hungarian
& $16.79\pm4.41$
& $19.23\pm4.63$
& $23.71\pm4.85$
& $\mathbf{28.53\pm5.34}$ \\

& Partial OT
& $22.95\pm3.91$
& $23.54\pm3.89$
& $23.71\pm4.16$
& $23.60\pm4.44$ \\
\midrule

\multirow{6}{*}{BrainHIVE}
& CSLS
& $\mathbf{29.16\pm6.47}$
& $29.59\pm6.91$
& $29.85\pm7.24$
& $30.22\pm7.41$ \\

& IS
& $19.96\pm3.53$
& $26.30\pm5.41$
& $31.59\pm7.02$
& $33.65\pm8.18$ \\

& GC
& $\underline{24.70\pm4.85}$
& $26.29\pm5.54$
& $28.72\pm5.85$
& $30.15\pm6.15$ \\

& Sinkhorn
& $23.27\pm3.35$
& $\mathbf{30.65\pm5.25}$
& $\mathbf{35.83\pm6.91}$
& $\mathbf{39.33\pm8.63}$ \\

& Hungarian
& $24.24\pm4.94$
& $26.11\pm5.56$
& $31.01\pm7.01$
& $35.67\pm8.30$ \\

& Partial OT
& $23.48\pm4.48$
& $\underline{30.21\pm5.53}$
& $\underline{35.27\pm7.34}$
& $\underline{37.95\pm8.49}$ \\
\bottomrule
\end{tabular*}
\end{center}
\end{table*}

\subsection{Statistical Significance of Reduced-Context Gains}
\label{app:statistical_significance}
To assess whether the reduced-context gains are consistent across held-out subjects, we conduct a subject-level significance analysis of CM-Top1 under the LOSO protocol. The ten held-out subjects serve as the units of analysis.
Let $m=\{25,50,100,150\}$ denote the set of reduced query-context sizes. For each subject $s$ and inference method $\Phi$, we average the retrieval gain over the six EEG--image representations and all context sizes in
\begin{equation}
\Delta_{s,\Phi}
=
\frac{1}{6}
\sum_{b=1}^{6}
\frac{1}{4}
\sum
\left[
\mathrm{CM\text{-}Top1}_{s,\Phi}^{(b)}(m)
-
\mathrm{Top1}_{s,\mathrm{ind}}^{(b)}
\right],
\label{eq:subject_level_gain}
\end{equation}
where $\mathrm{Top1}_{s,\mathrm{ind}}^{(b)}$ denotes the corresponding
independent-retrieval performance for subject $s$ on representation $b$. This produces one aggregated gain for each subject and inference method.

We report the mean subject-level gain together with its 95\% bootstrap confidence interval. To test whether the observed improvement differs significantly from zero, we apply a Wilcoxon signed-rank test to the ten subject-level gains. Since six structural inference methods are tested simultaneously, the resulting $p$-values are corrected using the Holm procedure.

\begin{table}[t]
\begin{center}
\caption{
Statistical significance of reduced-context CM-Top1 gains over independent retrieval.
Each comparison uses the ten held-out subjects as independent observations after averaging
over six representations and four reduced query context sizes. Reported $p$-values are
Holm-adjusted across the six inference methods.
}
\label{tab:significance_cm_top1}

\renewcommand{\arraystretch}{1.1}

\begin{tabular*}{\linewidth}{@{\extracolsep{\fill}}lccc@{}}
\toprule
Method & Mean SE (pp) & 95\% bootstrap CI & $p_{\mathrm{Holm}}$ \\
\midrule
CSLS       & +5.14 & [4.32, 5.92] & 0.0117 \\
IS         & +6.39 & [5.08, 7.56] & 0.0117 \\
GC         & +4.61 & [3.74, 5.45] & 0.0117 \\
Sinkhorn   & +7.99 & [6.38, 9.45] & 0.0117 \\
Hungarian  & +5.32 & [4.29, 6.33] & 0.0117 \\
Partial OT & +6.85 & [5.32, 8.12] & 0.0117 \\
\bottomrule
\end{tabular*}
\end{center}
\end{table}

As shown in Table~\ref{tab:significance_cm_top1}, all six structural inference methods achieve positive CM-Top1 gains over independent retrieval. Importantly,
the 95\% bootstrap confidence intervals of all methods remain entirely above zero. The improvements also remain statistically significant after correcting
for the six simultaneous comparisons, with
$p_{\mathrm{Holm}}=0.0117$ for every method. Sinkhorn exhibits the largest mean subject-level improvement, followed by Partial OT and IS, while even the
smallest average gain, obtained by GC, remains statistically reliable. These results provide subject-level evidence that the reduced-context gains reported in the main experiments are not attributable to a small subset of held-out subjects. Rather, the benefit of structural inference persists across held-out subjects under the LOSO protocol, with each held-out subject treated as an independent statistical unit.

\subsection{Context-Marginalized Top-5 Results}
\label{app:Context_Marginalized_context_top5}

\begin{table*}[t]
\caption{CM-Top5 structural exploitability (SE, pp), averaged over the reduced query-context sizes.
}
\label{tab:cm_top5_gain_backbones}
\begin{center}
\renewcommand{\arraystretch}{0.95}

\begin{tabular*}{\textwidth}{@{\extracolsep{\fill}}llccccccc@{}}
\toprule
Type & Method
& NICE
& ATM
& UBP
& BLUR
& HyFI
& BrainHIVE
& Average \\
\midrule

\multirow{3}{*}{Set-level}
& CSLS
& 2.43
& 4.98
& 10.81
& 9.24
& 14.01
& 10.95
& 8.74 \\

& IS
& 2.80
& 6.02
& \textbf{16.56}
& 9.24
& \textbf{20.53}
& 14.01
& 11.53 \\

& GC
& 2.27
& 6.75
& 15.15
& 6.71
& 18.67
& 12.61
& 10.36 \\
\midrule

\multirow{3}{*}{Assignment}
& Sinkhorn
& \textbf{5.55}
& \underline{7.41}
& \underline{15.52}
& \textbf{15.66}
& \underline{19.80}
& \textbf{18.45}
& \textbf{13.73} \\

& Hungarian
& --
& --
& --
& --
& --
& --
& -- \\

& Partial OT
& \underline{4.43}
& \textbf{7.52}
& 11.85
& \underline{12.58}
& 16.91
& \underline{17.41}
& \underline{11.78} \\

\bottomrule
\end{tabular*}
\end{center}
\end{table*}
Table~\ref{tab:cm_top5_gain_backbones} shows that the benefit of structural inference extends beyond Top-1 retrieval. Sinkhorn achieves the largest average CM-Top5 gain across representations (+13.73 pp), followed by Partial OT (+11.78 pp) and IS (+11.53 pp). The preferred method nevertheless varies across representations: Sinkhorn performs best on NICE, BLUR, and BrainHIVE, Partial OT on ATM, and IS on UBP and HyFI. These results indicate that structural inference can improve the broader candidate ranking, while the magnitude of the gain remains strongly dependent on the underlying EEG--image similarity space.

\section{Additional Analysis of Context Consistency}
\label{app:consistency_analysis}
\subsection{Detailed Context Consistency Results}
\label{app:detailed_cc}

Table~\ref{tab:detailed_cc} reports the complete same-query Context Consistency (CC) results for each reduced query context size $m\in\{25,50,100,150\}$. CSLS exhibits the strongest overall prediction stability, achieving the highest CC for nearly all representation--context combinations. Its consistency generally increases as more companion queries become available, for example from 76.92\% to 90.07\% on BLUR and from 81.21\% to 92.92\% on BrainHIVE as $m$ increases from 25 to 150. Partial OT is frequently the second-most consistent method and becomes the best-performing method on HyFI at $m=150$ (91.70\%), slightly exceeding CSLS (91.67\%). IS and Sinkhorn also show substantial increases in CC with larger query contexts on most representations.

Hungarian matching exhibits a qualitatively different context-response pattern. Rather than becoming consistently more stable as $m$ increases, its CC often decreases at intermediate or large context sizes. On BLUR, for example, CC drops from 76.69\% at $m=25$ to 63.44\% at $m=100$ and recovers only slightly to 65.12\% at $m=150$. Similar non-monotonic behavior is observed on UBP and HyFI. GC also shows non-monotonic consistency on several representations, although the variation is generally less pronounced. These results show that the context-averaged CC values reported in the main paper can conceal markedly different stability trajectories across individual context sizes.

\begin{table*}[t]
\centering
\caption{
Detailed Context Consistency (CC, \%) under reduced query
contexts $m$. Each entry reports mean$\pm$standard deviation across ten LOSO folds. 
}
\label{tab:detailed_cc}
\begin{center}

\setlength{\tabcolsep}{4.0pt}
\renewcommand{\arraystretch}{1.00}

\begin{tabular*}{\textwidth}{
@{\extracolsep{\fill}}
llcccc
@{}}
\toprule
Representation & Method
& $m=25$
& $m=50$
& $m=100$
& $m=150$ \\
\midrule

\multirow{6}{*}{NICE}
& CSLS
& $\mathbf{76.95\pm2.08}$
& $\mathbf{81.30\pm1.87}$
& $\mathbf{86.36\pm1.10}$
& $\mathbf{90.61\pm1.03}$ \\

& IS
& $35.96\pm0.85$
& $46.01\pm1.23$
& $61.47\pm1.40$
& $75.67\pm1.30$ \\

& GC
& $63.04\pm3.10$
& $58.07\pm2.23$
& $64.25\pm2.52$
& $75.74\pm1.08$ \\

& Sinkhorn
& $37.42\pm1.36$
& $50.01\pm1.89$
& $65.68\pm1.04$
& $79.16\pm1.30$ \\

& Hungarian
& $\underline{73.29\pm4.58}$
& $\underline{64.66\pm4.33}$
& $60.57\pm2.57$
& $64.26\pm1.82$ \\

& Partial OT
& $40.60\pm1.14$
& $53.12\pm1.97$
& $\underline{68.93\pm1.58}$
& $\underline{80.94\pm1.40}$ \\
\midrule

\multirow{6}{*}{ATM}
& CSLS
& $\mathbf{78.90\pm1.32}$
& $\mathbf{82.25\pm1.43}$
& $\mathbf{86.89\pm0.90}$
& $\mathbf{91.19\pm0.74}$ \\

& IS
& $31.32\pm1.38$
& $38.84\pm1.75$
& $51.58\pm1.83$
& $67.54\pm1.46$ \\

& GC
& $\underline{76.85\pm2.09}$
& $\underline{73.90\pm1.98}$
& $\underline{76.07\pm1.70}$
& $\underline{82.16\pm1.40}$ \\

& Sinkhorn
& $25.35\pm1.85$
& $36.08\pm1.30$
& $48.46\pm1.07$
& $63.48\pm1.03$ \\

& Hungarian
& $71.17\pm3.10$
& $62.87\pm3.03$
& $59.87\pm1.54$
& $64.72\pm0.82$ \\

& Partial OT
& $28.98\pm1.61$
& $38.55\pm1.25$
& $51.36\pm1.17$
& $66.23\pm1.15$ \\
\midrule

\multirow{6}{*}{UBP}
& CSLS
& $\mathbf{80.40\pm2.43}$
& $\mathbf{83.39\pm1.78}$
& $\mathbf{87.29\pm1.36}$
& $\mathbf{92.08\pm0.59}$ \\

& IS
& $61.17\pm2.56$
& $71.94\pm2.11$
& $81.89\pm1.76$
& $89.90\pm1.38$ \\

& GC
& $56.80\pm2.80$
& $55.10\pm3.14$
& $63.70\pm2.20$
& $76.67\pm1.05$ \\

& Sinkhorn
& $63.27\pm2.65$
& $73.25\pm2.51$
& $82.68\pm2.56$
& $90.09\pm1.30$ \\

& Hungarian
& $65.90\pm2.10$
& $60.31\pm2.84$
& $58.31\pm2.53$
& $63.61\pm1.29$ \\

& Partial OT
& $\underline{68.77\pm1.86}$
& $\underline{77.77\pm1.36}$
& $\underline{85.61\pm2.17}$
& $\underline{92.01\pm0.81}$ \\
\midrule

\multirow{6}{*}{BLUR}
& CSLS
& $\mathbf{76.92\pm1.79}$
& $\mathbf{80.53\pm1.32}$
& $\mathbf{85.02\pm2.15}$
& $\mathbf{90.07\pm1.83}$ \\

& IS
& $49.69\pm3.28$
& $61.81\pm3.68$
& $75.07\pm3.57$
& $85.37\pm2.50$ \\

& GC
& $55.89\pm3.15$
& $58.75\pm2.27$
& $68.58\pm1.47$
& $80.45\pm1.17$ \\

& Sinkhorn
& $55.19\pm2.48$
& $67.29\pm2.56$
& $79.34\pm1.56$
& $87.85\pm1.36$ \\

& Hungarian
& $\underline{76.69\pm6.09}$
& $67.84\pm4.29$
& $63.44\pm3.40$
& $65.12\pm1.74$ \\

& Partial OT
& $61.53\pm2.19$
& $\underline{72.15\pm2.22}$
& $\underline{82.81\pm1.77}$
& $\underline{89.57\pm1.36}$ \\
\midrule

\multirow{6}{*}{HyFI}
& CSLS
& $\mathbf{80.34\pm1.98}$
& $\mathbf{83.67\pm1.88}$
& $\mathbf{87.45\pm1.35}$
& $\underline{91.67\pm0.76}$ \\

& IS
& $59.20\pm1.89$
& $69.87\pm2.09$
& $81.03\pm1.68$
& $88.78\pm1.49$ \\

& GC
& $61.56\pm3.34$
& $59.01\pm2.59$
& $66.91\pm1.61$
& $77.90\pm1.40$ \\

& Sinkhorn
& $60.56\pm2.09$
& $70.74\pm1.93$
& $81.41\pm1.89$
& $89.86\pm1.22$ \\

& Hungarian
& $\underline{69.50\pm3.54}$
& $62.57\pm3.78$
& $59.71\pm2.18$
& $64.25\pm1.79$ \\

& Partial OT
& $67.18\pm1.75$
& $\underline{76.59\pm1.87}$
& $\underline{85.30\pm1.63}$
& $\mathbf{91.70\pm1.10}$ \\
\midrule

\multirow{6}{*}{BrainHIVE}
& CSLS
& $\mathbf{81.21\pm2.02}$
& $\mathbf{84.33\pm1.53}$
& $\mathbf{89.09\pm1.35}$
& $\mathbf{92.92\pm0.77}$ \\

& IS
& $31.67\pm1.28$
& $40.18\pm1.75$
& $54.71\pm2.38$
& $70.12\pm2.40$ \\

& GC
& $\underline{76.75\pm1.95}$
& $\underline{74.57\pm1.98}$
& $\underline{78.56\pm1.67}$
& $\underline{85.38\pm1.40}$ \\

& Sinkhorn
& $26.18\pm1.59$
& $38.57\pm1.41$
& $51.06\pm1.85$
& $65.14\pm1.50$ \\

& Hungarian
& $74.52\pm4.95$
& $66.08\pm3.90$
& $62.24\pm2.43$
& $65.86\pm2.09$ \\

& Partial OT
& $29.71\pm2.47$
& $41.30\pm2.65$
& $54.22\pm2.63$
& $68.57\pm1.88$ \\
\bottomrule
\end{tabular*}
\end{center}
\end{table*}

\subsection{Decomposing Context Consistency into Prediction Outcomes}
\label{app:cc_decomposition}

\begin{figure*}[t]
    \centering
    \includegraphics[
        width=\textwidth,
        keepaspectratio
    ]{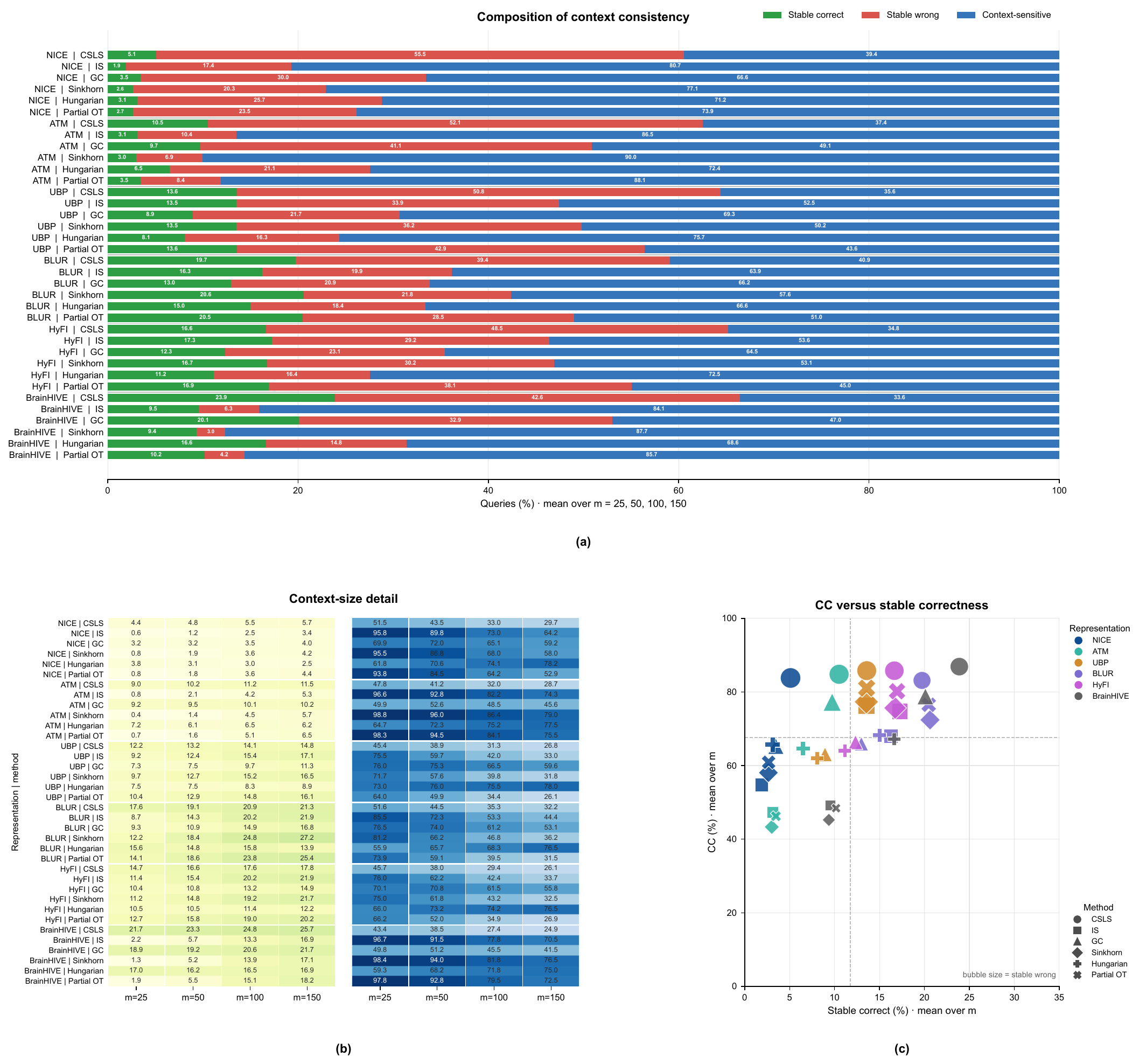}

    \caption{
Decomposition of same-query Context Consistency.
Queries are categorized as Stable Correct, Stable Wrong,
or Context-Sensitive according to their CM-Top1 predictions across sampled query contexts. (a) Average proportions of the three categories across $m$. (b) Stable-Correct and Context-Sensitive proportions at each context size. (c) Average CC versus Stable-Correct rate, where color denotes the representation, marker shape denotes the inference method, and bubble size represents the Stable-Wrong rate.
}
    \label{fig:stability_decomposition}
\end{figure*}

CC measures agreement among predictions for the same query across sampled contexts, but does not indicate whether a consistent prediction is correct. We therefore complement CC with a query-level decomposition of prediction stability. For each target query $q_i$, evaluated across $R_m$ sampled contexts, we assign its predictions to one of three mutually exclusive categories. A query is Stable Correct if all $R_m$ predictions identify its ground-truth image. It is Stable Wrong if all predictions identify the same non-ground-truth candidate, and Context-Sensitive otherwise. These categories are mutually exclusive and collectively exhaustive.

Figure~\ref{fig:stability_decomposition}(a) shows the resulting composition, averaged across four reduced context sizes and ten held-out subjects under the LOSO protocol. High prediction consistency does not necessarily imply stable correctness. CSLS, for example, contains a large Stable-Wrong component across several representations despite achieving the highest average CC in the main experiments. Its Stable-Wrong proportion reaches 55.5\% on NICE, 52.1\% on ATM, and 50.8\% on UBP, while the corresponding Stable-Correct proportions are substantially lower. Thus, much of the apparent stability of CSLS arises from repeatedly selecting the same incorrect candidate rather than consistently retrieving the ground truth.

Figure~\ref{fig:stability_decomposition}(b) further resolves these components by context size. For CSLS, IS, Sinkhorn, and Partial OT, increasing $m$ generally raises the Stable-Correct proportion and reduces the Context-Sensitive proportion. Hungarian exhibits a different pattern: its Context-Sensitive proportion often increases with $m$, without a corresponding improvement in Stable-Correct predictions. This pattern is consistent with the non-monotonic or decreasing CC profiles of Hungarian observed in the main experiments.

Figure~\ref{fig:stability_decomposition}(c)  relates average CC to the Stable-Correct proportion across representation–method combinations. Bubble size represents the Stable-Wrong proportion, allowing consistency, stable correctness, and persistent error to be examined jointly. The preferred operating region is the upper right with small bubbles, indicating high CC, a high Stable-Correct proportion, and little persistent error. Conversely, high CC combined with a low Stable-Correct proportion and large bubbles indicates repeated selection of the same incorrect prediction.

These results show that CC alone does not reveal the source of prediction stability across query contexts. Decomposing predictions into Stable-Correct, Stable-Wrong, and Context-Sensitive categories distinguishes consistent success, persistent error, and sensitivity to query composition.

\section{CMRE Convergence and Sampling Robustness}
\label{app:sampling_convergence}
\begin{figure*}[t]
    \centering
    \includegraphics[width=\textwidth]{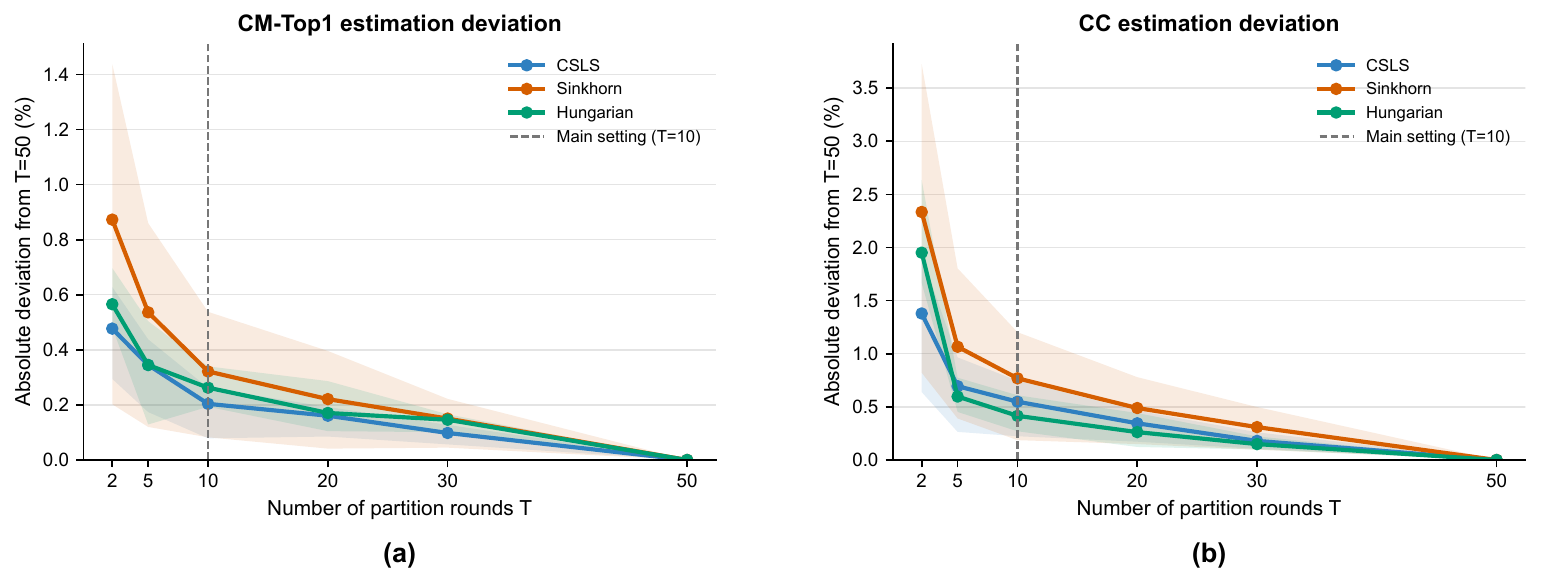}
   \caption{CMRE convergence over context-construction rounds on BLUR. Absolute deviations of (a) CM-Top1 and (b) Context Consistency (CC) from the $T=50$ reference as the number of partition rounds increases. Curves correspond to CSLS, Sinkhorn, and Hungarian and are averaged across context sizes. The dashed line marks the $T=10$ setting used in the main experiments. Both metrics are already close to the high-budget reference at $T=10$.
}
    \label{fig:cmre_sampling_convergence}
\end{figure*}
We conduct the convergence analysis on BLUR using three representative inference rules with distinct context-response characteristics: CSLS, which exhibits consistently high Context Consistency; Sinkhorn, which achieves the largest average reduced-context retrieval gain; and Hungarian, which shows pronounced context-dependent variation in both retrieval performance and prediction consistency. For each reduced context size $m\in\{25,50,100,150\}$, we generate a fixed schedule of 50 random partition rounds and evaluate progressively larger prefixes with $T\in\{2,5,10,20,30,50\}$, where $T$ denotes the number of partition rounds. The estimate obtained with $T=50$ is treated as a high-budget reference.

For inference method $\Phi$, we measure the average absolute deviation of CM-Top1 from the reference as
\begin{equation}
D_{\mathrm{Top1}}^{\Phi}(T)
=
\frac{1}{4}
\sum
\left|
\mathrm{CM\text{-}Top1}_{\Phi,T}(m)
-
\mathrm{CM\text{-}Top1}_{\Phi,50}(m)
\right|,
\label{eq:cmre_top1_convergence}
\end{equation}
and analogously define the deviation of Context Consistency as
\begin{equation}
D_{\mathrm{CC}}^{\Phi}(T)
=
\frac{1}{4}
\sum
\left|
\mathrm{CC}_{\Phi,T}(m)
-
\mathrm{CC}_{\Phi,50}(m)
\right|.
\label{eq:cmre_cc_convergence}
\end{equation}

Figure~\ref{fig:cmre_sampling_convergence} shows that both estimates rapidly approach the high-budget reference as the number of partition rounds increases. At the $T=10$ setting used in the main experiments, the average CM-Top1 deviations are only 0.204, 0.322, and 0.263 pp for CSLS, Sinkhorn, and Hungarian, respectively. The corresponding CC deviations are 0.549, 0.769, and 0.415 pp. Thus, even for inference rules with substantially different context-response behaviors, the estimates obtained with 10 partition rounds are already close to those obtained with 50 rounds. These results support the partition-round budget used in the main experiments. Increasing the number of rounds beyond $T=10$ produces only small changes in the reported CMRE estimates. This analysis evaluates prefix convergence within a shared 50-round partition schedule, rather than variability across independently resampled schedules.

To examine robustness to context-sampling schedules, we independently sample 10 contexts per query at each $m\in\{25,50,100,150\}$ using ten independent seeds. As shown in Table~\ref{tab:independent_sampling}, CM-Top1 and CC differ from the original results by at most 0.13 and 0.89 pp, respectively, with unchanged accuracy rankings among the three evaluated rules. The stable-prediction decompositions also remain similar at $m\in\{25,50,100\}$, with a maximum difference of 1.85 pp. At $m=150$, we use a matched-budget reference obtained by subsampling the original 30 predictions to 10 evaluations per query; the three decomposition categories then differ by at most 0.63 pp. These results indicate that the main retrieval accuracy rankings, context-consistency patterns, and stable-prediction decompositions are largely preserved under independently sampled contexts with matched evaluation budgets.

\begin{table}[t]
\centering
\caption{Comparison of the original CMRE protocol and independently sampled contexts on BLUR. Each entry reports reference $\rightarrow$ independent-sampling results (\%). Independent sampling uses ten contexts per query and ten seeds. At $m=150$, the reference for the three stable-prediction categories uses a matched budget of ten evaluations per query.}
\label{tab:independent_sampling}
\scriptsize
\setlength{\tabcolsep}{3pt}
\resizebox{\linewidth}{!}{
\begin{tabular}{llccccc}
\toprule
$m$ & Method & CM-Top1 & CC & Stable Correct & Stable Wrong & Context-sensitive \\
\midrule
25 & CSLS      & 26.13 $\to$ 26.02 & 76.92 $\to$ 76.76 & 17.65 $\to$ 17.48 & 30.70 $\to$ 30.93 & 51.65 $\to$ 51.59 \\
25 & Sinkhorn  & 30.13 $\to$ 30.05 & 55.19 $\to$ 54.30 & 12.15 $\to$ 11.80 & 6.70 $\to$ 6.76 & 81.15 $\to$ 81.45 \\
25 & Hungarian & 22.02 $\to$ 22.12 & 76.69 $\to$ 76.25 & 15.55 $\to$ 15.52 & 28.60 $\to$ 28.01 & 55.85 $\to$ 56.47 \\
\midrule
50 & CSLS      & 26.43 $\to$ 26.30 & 80.53 $\to$ 80.72 & 19.10 $\to$ 18.85 & 36.40 $\to$ 36.63 & 44.50 $\to$ 44.52 \\
50 & Sinkhorn  & 32.54 $\to$ 32.47 & 67.29 $\to$ 67.39 & 18.35 $\to$ 18.43 & 15.40 $\to$ 15.60 & 66.25 $\to$ 65.97 \\
50 & Hungarian & 24.25 $\to$ 24.36 & 67.84 $\to$ 68.37 & 14.85 $\to$ 15.15 & 19.45 $\to$ 19.70 & 65.70 $\to$ 65.16 \\
\midrule
100 & CSLS      & 26.61 $\to$ 26.63 & 85.02 $\to$ 85.46 & 20.85 $\to$ 20.69 & 43.85 $\to$ 44.29 & 35.30 $\to$ 35.02 \\
100 & Sinkhorn  & 34.14 $\to$ 34.06 & 79.34 $\to$ 80.08 & 24.75 $\to$ 24.89 & 28.50 $\to$ 30.21 & 46.75 $\to$ 44.90 \\
100 & Hungarian & 28.43 $\to$ 28.49 & 63.44 $\to$ 63.25 & 15.75 $\to$ 15.57 & 15.90 $\to$ 14.84 & 68.35 $\to$ 69.60 \\
\midrule
150 & CSLS      & 26.75 $\to$ 26.82 & 90.07 $\to$ 90.08 & 22.52 $\to$ 22.49 & 52.55 $\to$ 52.45 & 24.93 $\to$ 25.07 \\
150 & Sinkhorn  & 34.66 $\to$ 34.65 & 87.85 $\to$ 88.10 & 28.87 $\to$ 28.73 & 41.72 $\to$ 42.17 & 29.41 $\to$ 29.11 \\
150 & Hungarian & 33.36 $\to$ 33.38 & 65.12 $\to$ 65.47 & 18.18 $\to$ 18.49 & 14.14 $\to$ 14.46 & 67.69 $\to$ 67.06 \\
\bottomrule
\end{tabular}
}
\end{table}

\section{Robustness to Gallery Expansion and Query-context Variation}

\label{app:gallery_expansion}
\newcommand{\accse}[2]{\ensuremath{#1_{#2}}}
\newcommand{\bestse}[2]{\ensuremath{\mathbf{#1}_{#2}}}

\begin{table}[t]
\centering

\caption{Joint effects of query-context size $m$ and gallery size $N_g$ on BLUR. Top-1 accuracy (\%) is reported with SE (pp) as a subscript. bold indicates the best rule.}

\label{tab:joint_blur}
\small
\setlength{\tabcolsep}{6pt}
\renewcommand{\arraystretch}{1.08}
\begin{tabular*}{0.88\linewidth}{@{\extracolsep{\fill}}llcccc@{}}
\toprule
$N_g$ & $m$ & CSLS & Sinkhorn & Hungarian & Partial OT \\
\midrule
\multirow{5}{*}{200}
& 25  & \accse{26.13}{+6.08} & \bestse{30.13}{+10.08} & \accse{22.02}{+1.97} & \accse{29.57}{+9.52} \\
& 50  & \accse{26.43}{+6.38} & \bestse{32.54}{+12.49} & \accse{24.25}{+4.20} & \accse{30.42}{+10.37} \\
& 100 & \accse{26.61}{+6.56} & \bestse{34.14}{+14.09} & \accse{28.43}{+8.38} & \accse{30.98}{+10.93} \\
& 150 & \accse{26.75}{+6.70} & \bestse{34.66}{+14.61} & \accse{33.36}{+13.31} & \accse{31.00}{+10.95} \\
& 200 & \accse{26.95}{+6.90} & \accse{35.25}{+15.20} & \bestse{40.15}{+20.10} & \accse{30.90}{+10.85} \\
\midrule
\multirow{5}{*}{500}
& 25  & \accse{15.49}{+2.63} & \accse{15.39}{+2.53} & \accse{13.57}{+0.71} & \bestse{16.59}{+3.73} \\
& 50  & \accse{15.80}{+2.94} & \accse{16.95}{+4.09} & \accse{14.27}{+1.41} & \bestse{17.75}{+4.89} \\
& 100 & \accse{16.04}{+3.18} & \accse{18.14}{+5.28} & \accse{15.68}{+2.82} & \bestse{18.45}{+5.59} \\
& 150 & \accse{16.05}{+3.19} & \accse{18.45}{+5.59} & \accse{16.85}{+3.99} & \bestse{18.70}{+5.84} \\
& 200 & \accse{16.03}{+3.17} & \accse{18.55}{+5.69} & \accse{17.97}{+5.11} & \bestse{18.88}{+6.02} \\
\midrule
\multirow{5}{*}{1000}
& 25  & \accse{10.50}{+1.78} & \accse{9.61}{+0.89} & \accse{9.09}{+0.37} & \bestse{10.75}{+2.03} \\
& 50  & \accse{10.69}{+1.97} & \accse{10.92}{+2.20} & \accse{9.41}{+0.69} & \bestse{11.81}{+3.09} \\
& 100 & \accse{10.88}{+2.16} & \accse{11.76}{+3.04} & \accse{9.98}{+1.26} & \bestse{12.37}{+3.65} \\
& 150 & \accse{10.94}{+2.22} & \accse{12.01}{+3.29} & \accse{10.45}{+1.73} & \bestse{12.64}{+3.92} \\
& 200 & \accse{10.90}{+2.18} & \accse{12.19}{+3.47} & \accse{10.91}{+2.19} & \bestse{12.76}{+4.04} \\
\midrule
\multirow{5}{*}{2000}
& 25  & \bestse{6.73}{+0.93} & \accse{5.67}{-0.13} & \accse{5.95}{+0.15} & \accse{6.48}{+0.68} \\
& 50  & \accse{6.83}{+1.03} & \accse{6.64}{+0.84} & \accse{6.15}{+0.35} & \bestse{7.39}{+1.59} \\
& 100 & \accse{7.09}{+1.29} & \accse{7.06}{+1.26} & \accse{6.43}{+0.63} & \bestse{7.96}{+2.16} \\
& 150 & \accse{7.11}{+1.31} & \accse{7.23}{+1.43} & \accse{6.62}{+0.82} & \bestse{8.12}{+2.32} \\
& 200 & \accse{7.05}{+1.25} & \accse{7.45}{+1.65} & \accse{6.70}{+0.90} & \bestse{8.05}{+2.25} \\
\bottomrule
\end{tabular*}
\end{table}

\begin{table}[t]
\centering

\caption{Joint effects of query-context size $m$ and gallery size $N_g$ on UBP. Top-1 accuracy (\%) is reported with SE (pp) as a subscript. bold indicates the best rule.}

\label{tab:joint_ubp}
\small
\setlength{\tabcolsep}{5pt}
\renewcommand{\arraystretch}{1.08}
\begin{tabular*}{0.88\linewidth}{@{\extracolsep{\fill}}llcccc@{}}
\toprule
$N_g$ & $m$ & CSLS & Sinkhorn & Hungarian & Partial OT \\
\midrule
\multirow{5}{*}{200}
& 25  & \accse{16.30}{+5.85} & \bestse{19.04}{+8.59} & \accse{13.95}{+3.50} & \accse{18.11}{+7.66} \\
& 50  & \accse{16.66}{+6.21} & \bestse{19.89}{+9.44} & \accse{15.57}{+5.12} & \accse{18.60}{+8.15} \\
& 100 & \accse{17.01}{+6.56} & \bestse{20.15}{+9.70} & \accse{18.48}{+8.03} & \accse{18.69}{+8.24} \\
& 150 & \accse{17.19}{+6.74} & \accse{20.22}{+9.77} & \bestse{20.91}{+10.46} & \accse{18.71}{+8.26} \\
& 200 & \accse{17.20}{+6.75} & \accse{20.15}{+9.70} & \bestse{23.60}{+13.15} & \accse{18.75}{+8.30} \\
\midrule
\multirow{5}{*}{500}
& 25  & \accse{8.42}{+2.97} & \bestse{9.53}{+4.08} & \accse{6.76}{+1.31} & \accse{9.36}{+3.91} \\
& 50  & \accse{8.80}{+3.35} & \bestse{10.09}{+4.64} & \accse{7.51}{+2.06} & \accse{9.65}{+4.20} \\
& 100 & \accse{9.20}{+3.75} & \bestse{10.25}{+4.80} & \accse{8.55}{+3.10} & \accse{9.66}{+4.21} \\
& 150 & \accse{9.27}{+3.82} & \bestse{10.34}{+4.89} & \accse{9.36}{+3.91} & \accse{9.62}{+4.17} \\
& 200 & \accse{9.29}{+3.84} & \bestse{10.39}{+4.94} & \accse{10.20}{+4.75} & \accse{9.57}{+4.12} \\
\midrule
\multirow{5}{*}{1000}
& 25  & \accse{4.99}{+1.78} & \accse{5.72}{+2.51} & \accse{3.81}{+0.60} & \bestse{5.81}{+2.60} \\
& 50  & \accse{5.24}{+2.03} & \bestse{6.09}{+2.88} & \accse{4.28}{+1.07} & \accse{5.98}{+2.77} \\
& 100 & \accse{5.49}{+2.28} & \bestse{6.18}{+2.97} & \accse{4.80}{+1.59} & \accse{5.94}{+2.73} \\
& 150 & \accse{5.50}{+2.29} & \bestse{6.26}{+3.05} & \accse{5.12}{+1.91} & \accse{5.90}{+2.69} \\
& 200 & \accse{5.60}{+2.39} & \bestse{6.29}{+3.08} & \accse{5.49}{+2.28} & \accse{5.88}{+2.67} \\
\midrule
\multirow{5}{*}{2000}
& 25  & \accse{2.66}{+0.81} & \accse{3.27}{+1.42} & \accse{2.04}{+0.19} & \bestse{3.48}{+1.63} \\
& 50  & \accse{2.77}{+0.92} & \accse{3.35}{+1.50} & \accse{2.27}{+0.42} & \bestse{3.45}{+1.60} \\
& 100 & \accse{2.89}{+1.04} & \accse{3.35}{+1.50} & \accse{2.60}{+0.75} & \bestse{3.41}{+1.56} \\
& 150 & \accse{2.85}{+1.00} & \accse{3.38}{+1.53} & \accse{2.89}{+1.04} & \bestse{3.41}{+1.56} \\
& 200 & \accse{2.85}{+1.00} & \bestse{3.35}{+1.50} & \accse{3.15}{+1.30} & \bestse{3.35}{+1.50} \\
\bottomrule
\end{tabular*}
\end{table}


The standard THINGS-EEG2 retrieval protocol uses a closed gallery with $N_q=N_g=200$, containing one ground-truth image for each held-out EEG query. To examine whether structural gains depend on this setting, we keep the 200 queries and their ground-truth images fixed while expanding the gallery to $N_g\in\{200,500,1000,2000\}$. We construct a fixed pool of 1,800 distractors from unused images in the original THINGS collection. We exclude all images used in the THINGS-EEG2 training and test sets and all images belonging to the 200 test concepts, ensuring that the distractors neither duplicate evaluated concepts nor have corresponding EEG queries. Expanded galleries are formed by adding subsets of this pool; at $N_g=2000$, all 1,800 distractors are included. The EEG and image representations remain frozen throughout, so only the test-time gallery composition changes.

\begin{figure*}[t]
    \centering
    \includegraphics[width=\textwidth]{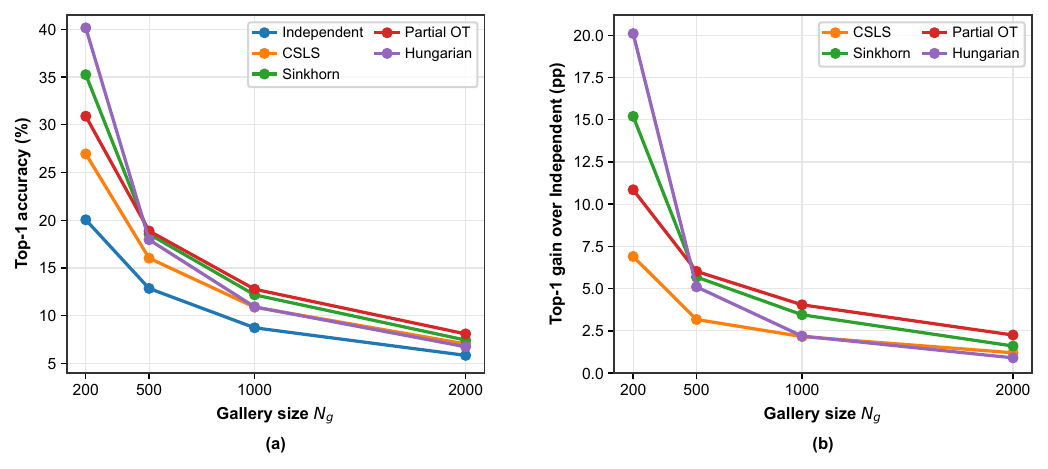}
  \caption{
Robustness to gallery expansion on BLUR.
(a) Top-1 retrieval accuracy as the gallery size increases from
$N_g=200$ to $N_g=2000$ while keeping all 200 EEG queries fixed.
(b) Corresponding Top-1 gain over independent retrieval.
}
    \label{fig:gallery_expansion}
\end{figure*}

Figure~\ref{fig:gallery_expansion} summarizes the BLUR results. As expected, absolute Top-1 accuracy decreases for every method as $N_g$ increases, reflecting greater retrieval difficulty (Figure 9(a)). Structure-aware inference nevertheless remains superior to independent retrieval across all expanded-gallery settings, although its gains decrease substantially with $N_g$. Hungarian achieves the largest gain in the original closed gallery, improving Top-1 accuracy by 20.10 percentage points at $N_g=200$. Its gain falls to 5.11, 2.19, and 0.90 percentage points at $N_g=500$, $1000$, and $2000$, respectively. By contrast, Partial OT retains the largest gain after expansion, achieving 6.02, 4.04, and 2.25 percentage points at the three larger gallery sizes, while Sinkhorn and CSLS show intermediate behavior.

These results indicate that the effectiveness of structural inference depends not only on the frozen similarity representation but also on the structural assumptions induced by the test gallery. Hard one-to-one assignment benefits strongly from the original closed-gallery setting, whereas partial transport is more robust when additional unmatched gallery candidates are introduced.

We further evaluate BLUR and UBP across five query-context sizes ($m\in\{25,50,100,150,200\}$) and four gallery sizes ($N_g\in\{200,500,1000,2000\}$), using CSLS, Sinkhorn, Hungarian, and Partial OT without retraining. Results are averaged over ten LOSO subjects. The independent-retrieval Top-1 accuracies are 20.05\%, 12.86\%, 8.72\%, and 5.80\% on BLUR, and 10.45\%, 5.45\%, 3.21\%, and 1.85\% on UBP, respectively, as $N_g$ increases. As shown in Tables~\ref{tab:joint_blur} and~\ref{tab:joint_ubp}, the preferred inference rule depends on both context size and gallery configuration. At $N_g=200$, Hungarian is the full-context winner on both representations but becomes suboptimal under several reduced contexts. In contrast, at $N_g=500$, Partial OT on BLUR and Sinkhorn on UBP remain optimal across all evaluated context sizes, yielding zero selection regret among the four compared rules. These findings show that full-context rule selection is not universally unreliable; its reliability varies with the query--gallery configuration. CMRE enables such configuration-dependent selection patterns to be systematically identified and quantified, beyond full-context accuracy alone.

\section{Discussion and Limitations}

CMRE characterizes how a fixed inference rule behaves when the query context available at test time changes. This perspective complements conventional full-context evaluation, but it does not model every plausible deployment setting. In particular, our contexts are generated through balanced random constructions. Real query streams may instead be temporally correlated, non-uniform, or task dependent. Although the balanced construction controls the number of appearances of each query, sampled contexts overlap and therefore do not constitute independent observations.

We intentionally keep each inference rule and its hyperparameters fixed across frozen representations and context sizes. This isolates the effect of available query context from context-specific retuning. Consequently, our results describe the behavior of the evaluated rules at shared fixed operating points, rather than their optimally tuned performance under every representation or context size. Moreover, Context Consistency measures agreement among predictions across contexts, not their correctness. A method can therefore be highly consistent while making consistently incorrect predictions.

Our empirical study focuses primarily on THINGS-EEG2. The complementary THINGS-MEG evaluation includes four participants and two frozen representations, and should therefore be interpreted as supporting evidence rather than a comprehensive cross-modal benchmark. The gallery-expansion analysis further shows that the gains of hard one-to-one assignment depend strongly on the closed, bijective gallery assumption. Its behavior is less favorable when unmatched distractors are introduced. Extending CMRE to larger cohorts, broader neural recording modalities, more realistic query streams, and non-bijective retrieval settings is an important direction for future work.
\end{document}

%% file: math_commands.tex
\usepackage{amsmath,amsfonts,bm}

\def\eqref#1{equation~\ref{#1}}

\def\1{\bm{1}}

\DeclareMathAlphabet{\mathsfit}{\encodingdefault}{\sfdefault}{m}{sl}
\SetMathAlphabet{\mathsfit}{bold}{\encodingdefault}{\sfdefault}{bx}{n}

